\documentclass[10pt,twocolumn,letterpaper]{article}

\usepackage[pagenumbers]{cvpr} 

\usepackage[dvipsnames]{xcolor}

\usepackage{subcaption}
\usepackage{caption}
\usepackage{newfloat}
\usepackage{listings}
\usepackage{xspace}

\usepackage{graphbox}
\usepackage{graphicx}
\usepackage{multirow}
\usepackage{wrapfig}
\usepackage{tikz}
\usepackage{float}

\usepackage[export]{adjustbox}
\usepackage{booktabs}
\usepackage{stackengine}
\usepackage{tablefootnote}
\usepackage{threeparttable}

\usepackage{amsmath}
\usepackage{amssymb}
\usepackage{bbm}
\usepackage{dsfont}
\usepackage{mathtools}

\usepackage{color}
\usepackage{colortbl}
\usepackage[normalem]{ulem}

\newcommand{\hfilll}{\hspace{0pt plus 1filll}}

\makeatletter
\newcommand*{\rom}[1]{\expandafter\@slowromancap\romannumeral #1@}
\makeatother

\makeatletter

\def\OurTitle{Aligning and Prompting Everything All at Once for Universal Visual Perception}
\def\OurMethodFullName{APE, a universal visual perception model for \textbf{a}ligning and \textbf{p}rompting \textbf{e}verything all at once in an image\xspace}
\def\OurMethod{APE\xspace}
\def\OurMethods{APE\xspace}

\makeatother

\newcommand{\checkmarknew}{\raisebox{0.6ex}{\scalebox{0.7}{$\sqrt{}$}}}
\newcommand{\crossmarknew}{\scalebox{0.85}[1]{$\times$}}

\usepackage[normalem]{ulem}

\newcommand{\veryshortarrow}[1][3pt]{\mathrel{%
   \hbox{\rule[\dimexpr\fontdimen22\textfont2-.2pt\relax]{#1}{.4pt}}%
   \mkern-4mu\hbox{\usefont{U}{lasy}{m}{n}\symbol{41}}}}

\makeatletter

\definecolor{cvprblue}{rgb}{0.21,0.49,0.74}
\usepackage[pagebackref,breaklinks,colorlinks,citecolor=cvprblue]{hyperref}

\def\paperID{1936} 
\def\confName{CVPR}
\def\confYear{2024}

\title{\OurTitle}

\author{
	Yunhang Shen$^1$, Chaoyou Fu$^1$, Peixian Chen$^1$, Mengdan Zhang$^1$ \\
	Ke Li$^1$, Xing Sun$^1$, Yunsheng Wu$^1$, Shaohui Lin$^2$, Rongrong Ji$^3$
	\\
	$^1$Tencent Youtu Lab \quad
	$^2$School of Computer Science and Technology \\ East China Normal University, Shanghai, China \quad
	$^3$Key Laboratory of Multimedia Trusted \\ Perception and Efficient Computing, Ministry of Education of China, Xiamen University, China \\
	{\tt\small \{shenyunhang01, bradyfu24\}@gmail.com, shlin@cs.ecnu.edu.cn, rrji@xmu.edu.cn} \\ 
	{\tt\small \{davinazhang, peixianchen, tristanli, winfredsun, simonwu\}@tencent.com}
}

\begin{document}
\maketitle

\begin{abstract}

Vision foundation models
have been explored recently to build general-purpose vision systems.
However, predominant paradigms, driven by casting instance-level tasks as an object-word alignment, bring heavy cross-modality interaction, which is not effective in prompting object detection and visual grounding.
Another line of work that focuses on pixel-level tasks often encounters a large annotation gap of things and stuff, and suffers from mutual interference between foreground-object and background-class segmentation.
In stark contrast to the prevailing methods, we present \OurMethodFullName to perform diverse tasks, \ie, detection, segmentation, and grounding, as an instance-level sentence-object matching paradigm.
Specifically, \OurMethod advances the convergence of detection and grounding by reformulating language-guided grounding as open-vocabulary detection, which efficiently scales up model prompting to thousands of category vocabularies and region descriptions while maintaining the effectiveness of cross-modality fusion.
To bridge the granularity gap of different pixel-level tasks, \OurMethod equalizes semantic and panoptic segmentation to proxy instance learning by considering any isolated regions as individual instances.
\OurMethod aligns vision and language representation on broad data with natural and challenging characteristics all at once without task-specific fine-tuning.
The extensive experiments on over $160$ datasets demonstrate that, with only one-suit of weights, \OurMethod outperforms (or is on par with) the state-of-the-art models, proving that an effective yet universal perception for anything aligning and prompting is indeed feasible.
Codes and trained models are released at \url{https://github.com/shenyunhang/APE}.

\end{abstract}
\section{Introduction}
\label{sec_Introduction}

\begin{figure}[t]
\begin{center}
\begin{subfigure}[t]{0.40\textwidth}
\begin{center}
\includegraphics[width=1.0\textwidth]{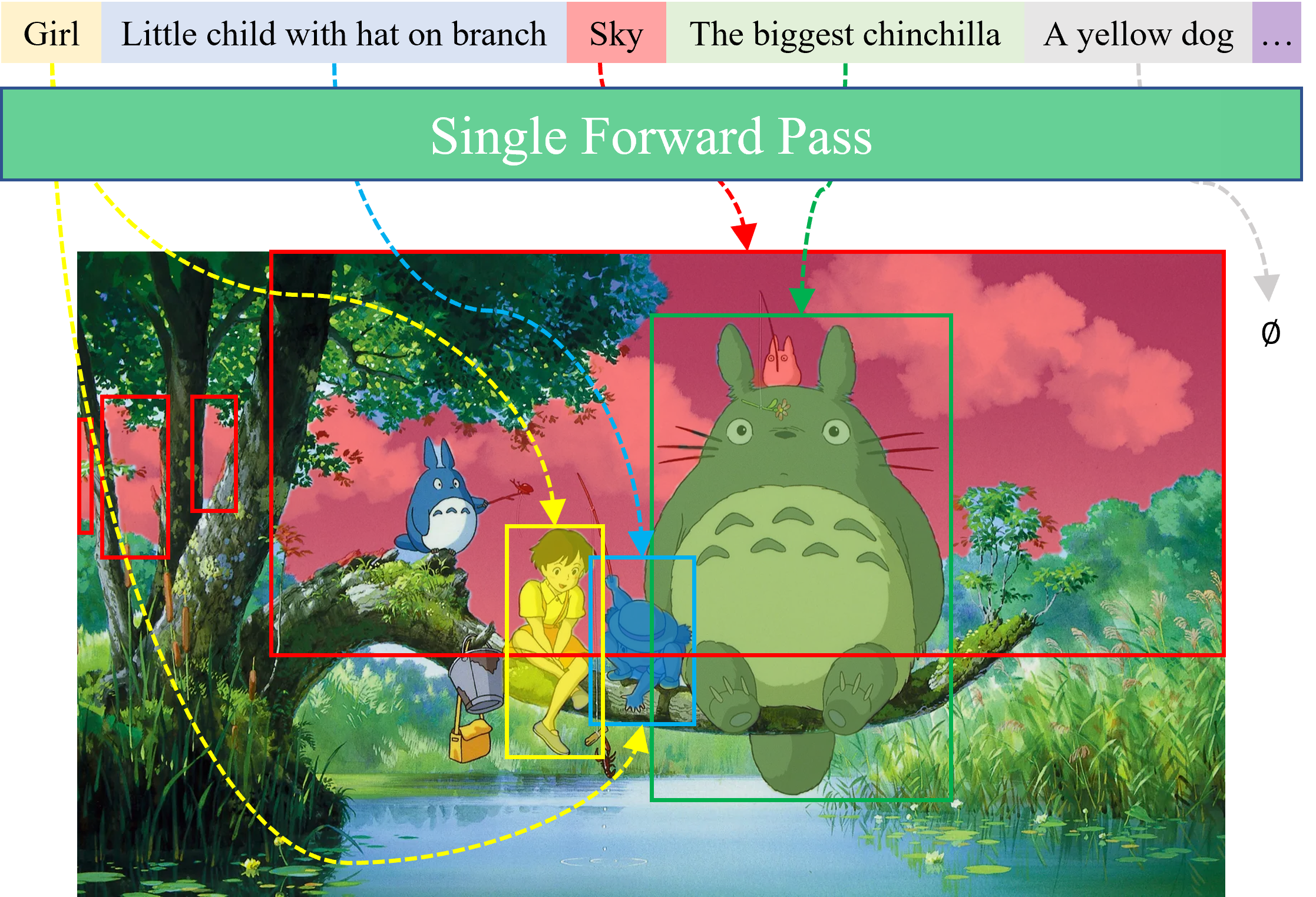}
\end{center}
\end{subfigure}
\begin{subfigure}[t]{0.34\textwidth}
\begin{center}
\includegraphics[width=1.0\textwidth]{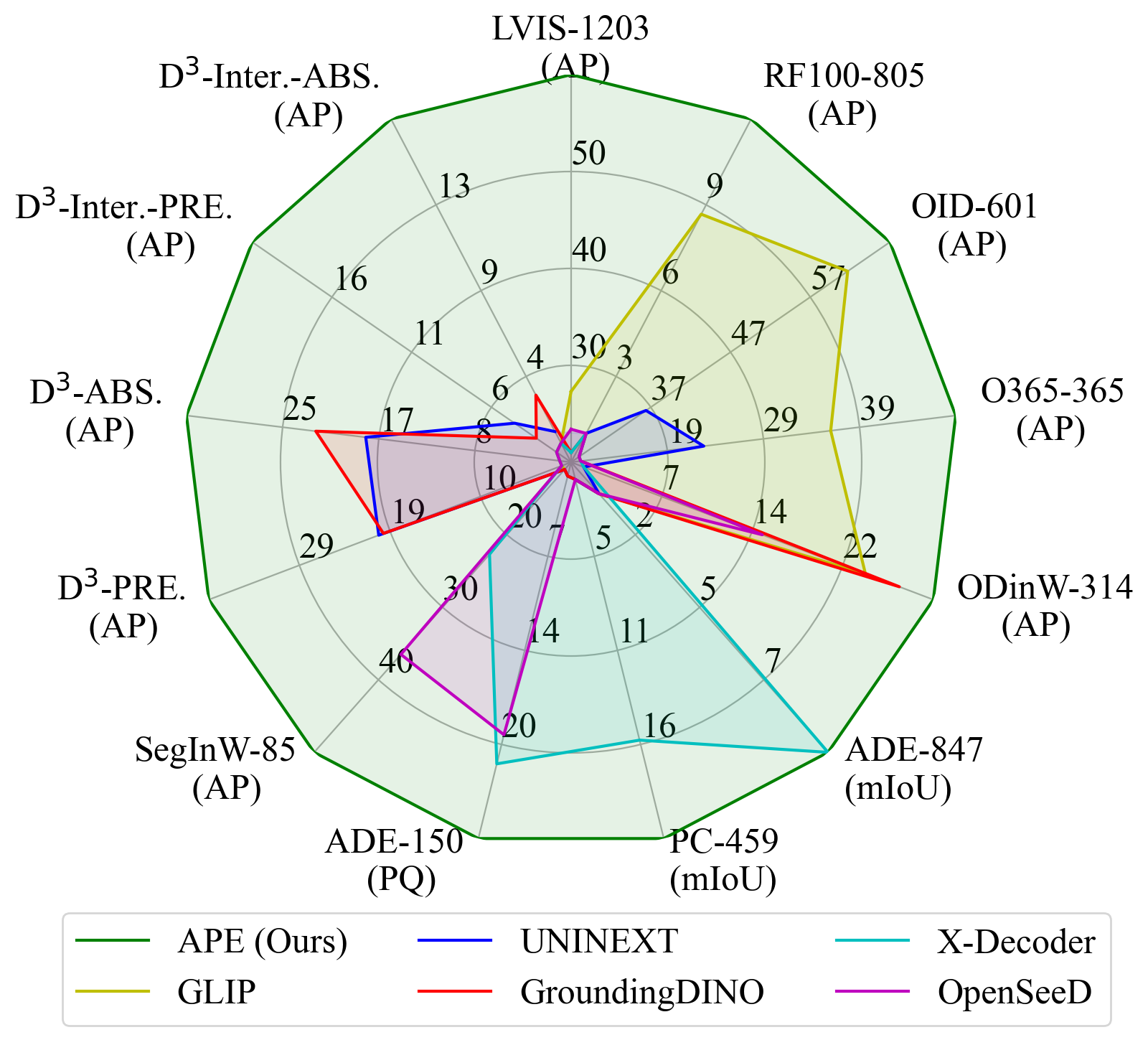}
\end{center}
\end{subfigure}
\end{center}
\caption{
\OurMethod supports prompting thousands of things, stuff, and sentences in a single forward pass and performing various segmentation without granularity discrepancy.
\OurMethod achieves new state-of-the-art or competitive performance on over $160$ datasets with one suite of parameters.
}
\label{figure_head}
\end{figure}

Developing vision systems that recognize and localize a wide range of basic concepts and can be transferable to
novel concepts or domains, has emerged as an important research topic in the community.
In light of the strong transferability demonstrated by LLMs, many researchers have attempted to build advanced vision foundation models~(VFMs) to serve general-purpose vision tasks.

Generally, the existent VFMs are roughly categorized into three groups.
The first one is to learn all-purposed visual features via self-supervised learning, \eg, DINO~\cite{DINO} and iBOT~\cite{iBOT}, and align text-image corpora with weakly-supervised learning, such as CLIP~\cite{CLIP} and ALIGN~\cite{ALIGN}.
Despite the promising feature transferability, the aforementioned methods often require individual adapters for downstream tasks.
The second group aligns region and text representation for instance perception tasks, such as GLIP~\cite{GLIP}, UNINEXT~\cite{UNINEXT}, and GroundingDINO~\cite{GroundingDINO}.
As they formulate this problem as a visual grounding problem with deep region-word fusion, it is incapable of operating on a large number of categories and grounding phrases at once~\cite{DetCLIP}.
The other group focuses on generic segmentation tasks, such as SAM~\cite{SAM}, X-Decoder~\cite{X-Decoder}, OpenSeeD~\cite{OpenSeeD}, and SEEM~\cite{SEEM}.
However, they usually suffer from the granularity discrepancy between foreground objects and background stuff, as foreground objects often perform object-level instance segmentation while background stuff corresponds to class-level semantic segmentation.
Previous methods~\cite{PanopticSegFormer,MaskDINO,OpenSeeD} decouple foreground and background learning with private queries and workflows, which involves manual prior knowledge to route each concept.

To develop VFMs that address the above problems, 
this work explores efficient promptable perception models and handles diverse semantic concepts for detection, foreground and background segmentation, and grounding.
To address the heavy computational cost of vision-language fusion, we aggregate compact sentence representation with gated cross-modality interaction and efficiently adapt the concept of vocabularies and sentences into a common embedding space.
To address the granularity discrepancy of foreground things and background stuff, we equalize their granularity by decomposing the category-level segmentation learning into the instance-level proxy objective, forming a single instance segmentation task.
Then instance-level patterns are ready to project back to category-level segments during inference.
By eliminating the discrepancy between foreground and background, it is granularity-friendly to learn from category-aware and category-agnostic segmentation data without distinguishing things and stuff manually.

To this end, we propose a \OurMethodFullName, which performs foundational vision tasks with an instance-level region-sentence interaction and matching paradigm.
We characterize several important capabilities that maximize \OurMethod's practicality in real-world scenarios from three perspectives:
(1) Task generalization:
\OurMethod is built based on DETR~\cite{DETR} framework to perform a wide array of semantic understanding tasks, which is capable of predicting labels, boxes, and masks for any object, region, and parts.
Specifically, we unify object detection of common and long-tailed vocabularies, image segmentation for various granularity, and visual grounding, into an instance-level detection transformer framework.
(2) Data diversity:
\OurMethod is trained on broad data sources all at once, ranging from long-tailed categories, federated annotations, anything segmentation, and hybrid vocabulary- and sentence-described concepts.
(3) Effective description prompting:
It is feasible to query \OurMethod with thousands of text prompts for object vocabularies and sentence descriptions, which aggregates the word-level prompt embedding for effective gated cross-modality fusion and alignment.

Benchmarked on over $160$ 
datasets,
\OurMethod achieves the state-of-the-art~(SotA) or competitive performance with one-suit of weights at various visual perception tasks, demonstrating the generalization and practicality of \OurMethod as VFMs.
We hope to facilitate the community on wide real-life applications.
Extensive ablation studies also verify the efficiency and effectiveness of each proposed component.

Conclusively, our contributions are the following:
\begin{itemize}
    \item 
    We present a \OurMethodFullName
    , which is trained on broad data at scale and provides SotA performance without task-specific fine-tuning.
    \item
    We reformulate the visual grounding as open-vocabulary detection with region-sentence vision-language interaction and matching, which significantly improves the efficiency of model querying for large-scale text prompts.
    \item
    We bridge the granularity gap of various segmentation patterns by transforming learning into the object-level proxy objective, which models thing and stuff categories equally without category-specific design.
\end{itemize}

\section{Related Work}
\label{sec_Related_Work}

\begin{figure*}[t]
\begin{center}
\begin{subfigure}[t]{1.0\textwidth}
\begin{center}
\includegraphics[width=1.0\textwidth]{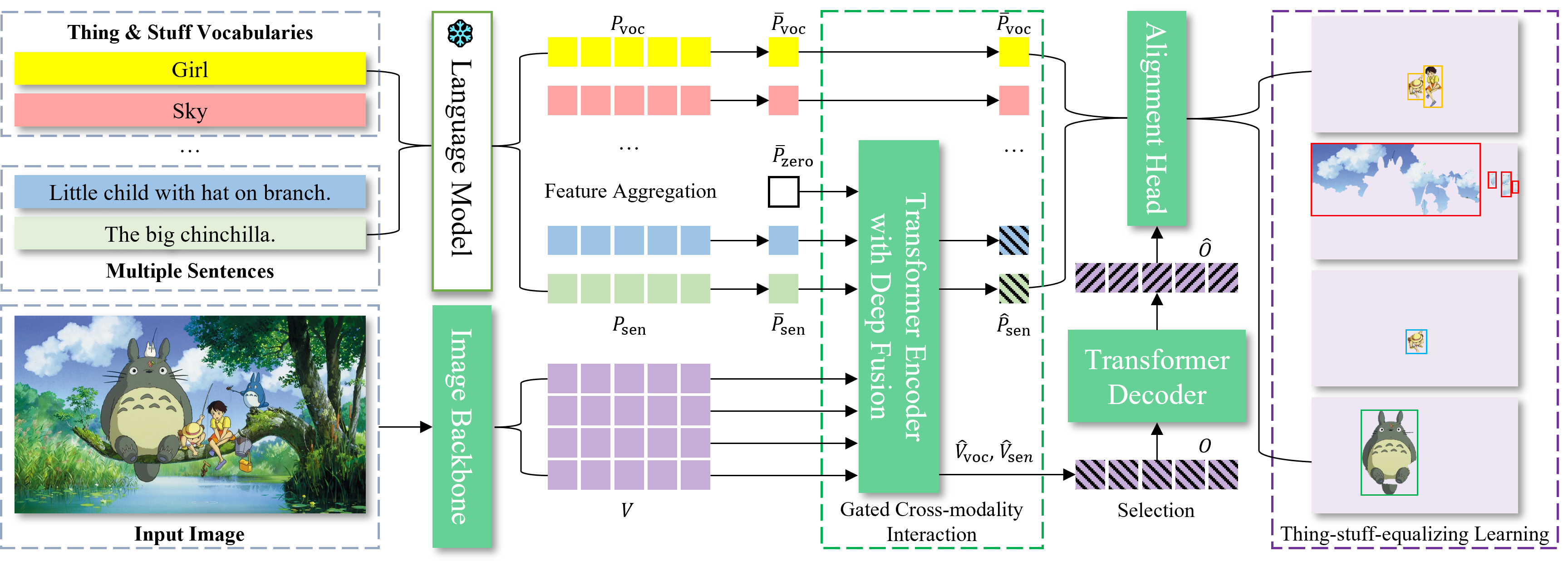}
\end{center}
\end{subfigure}
\end{center}
\caption{The overall framework of the proposed \OurMethod.
First, the image backbone and language model extract discrete visual embeddings $V$ and text embeddings $P$ for a given image and corresponding text prompts, respectively.
Second, the word-level text embedding $P$ is further aggregated into sentence-level embeddings $\bar{P}$.
Then, the cross-modality encoder fuses information from two modalities to condition the object queries $O$ on text queries and update text embeddings $\hat{P}$.
A transformer decoder generated final object embeddings $\hat{O}$ from object queries $O$.
Finally, a visual-language alignment module to predict the correct pairings of regions and prompts.
}
\label{figure_framework}
\end{figure*}

Unified vision-language models have recently drawn a lot of attention because of their great flexibility in generalizing to various tasks.
Mainstream approaches can be roughly categorized into two main types of goals, namely, ground anything and segment anything.
The former unified instance-level tasks, such as object detection and visual grounding, as region-word grounding learning, while the latter focuses on promptable and interactive learning for pixel-level segmentation with dense outputs.

\subsection{Unified Detection and Grounding}
Pix2Seq~\cite{Pix2Seq} and SeqTR~\cite{SeqTR} design a pixel-to-sequence interface for various tasks, such as object detection, instance segmentation, and visual grounding.
MDETR~\cite{MDETR} links the output tokens of DETR to specific words with region-word alignment.
GLIP~\cite{GLIP}
formulates detection as grounding and learns instance-level visual representation with language-aware deep fusion.
UNINEXT~\cite{UNINEXT} further supports prompts in both text and image for instance-level, \ie, foreground objects, perception tasks.
GroundingDINO~\cite{GroundingDINO} introduces additional cross-modality to the encoder, query selection, and decoder of detection transformers.

While promising generalization performance is presented, we argue that casting detection as a grounding problem via object-word fusion and alignment leads to inefficient interaction between vision and language.
Specifically, they can not prompt a large number of vocabularies or expressions in one forward due to the limit of GPU memory footprint and token length of text models.
For example, LVIS~\cite{LVIS} and D$^3$~\cite{DOD} have $1203$ vocabularies and $422$ descriptions, on which the previous models~\cite{GLIP,UNINEXT,GroundingDINO} require about $30$ and $422$ forwards to infer on a single image.
To address this drawback, we reformulate detection and grounding tasks as instance-level region-sentence matching, which is feasible to query models with thousands of concepts at scale.
Meanwhile, \OurMethod also avoids the additional processing of extracting the root object in given sentences.

\subsection{Unified Image Segmentation}
Mask2Former~\cite{Mask2Former} and MaskDINO~\cite{MaskDINO} present a universal architecture capable of handling semantic, instance, and panoptic segmentation for close-set categories.
ODISE~\cite{ODISE} leverages the frozen internal representation of text-to-image diffusion models for open-vocabulary panoptic segmentation.
X-Decoder~\cite{X-Decoder} and SEEM~\cite{SEEM} introduce a query-based segmentation architecture to support generic, referring, and interactive segmentation, and image-level vision-language understanding tasks.
However, they suffer from the mutual interference between things and stuff within each query.
To alleviate granularity discrepancy, OpenSeeD~\cite{OpenSeeD} and HIPIE~\cite{HIPIE} decouple foreground things and background stuff with separate decoders instead of one unified one.

However, decoupling learning~\cite{OpenSeeD,HIPIE} requires manually defining categories into things and stuff for both training and inference, which does not apply to segmentation data without semantic labels, such as SA-1B~\cite{SAM}.
In this paper, we formulate foreground and background equally with the proxy instance-level objective.
And the instance-level outputs are ready to convert to segmentation predictions of different formats to satisfy the desired granularity of tasks, \ie, semantic and panoptic segmentation.

\section{Method}

As shown in Fig.~\ref{figure_framework}, \OurMethod consists of a vision backbone for image feature extraction, a language model for text feature extraction, a transformer encoder with cross-modality deep fusion following GLIP~\cite{GLIP}, and a transformer decoder.
\OurMethod is expected to output a set of scores, boxes, and masks for a given image and a set of prompts, which could contain a large number of (thing and stuff) vocabularies and sentences.
To achieve this, we first construct compact text embeddings to efficiently perform vision-language interaction~(Sec.~\ref{sec_prompt}), which allows prompting \OurMethod with large-scale concepts in a single forward pass.
Then, we decompose the class-level learning into the object-level proxy objective~(Sec.~\ref{sec_thing_stuff}), which equalizes the granularity of foreground things and background stuff.
Finally, we assemble the publicly available data of detection, segmentation, and grounding to exclusively align vision-language~(Sec.~\ref{sec_Aligning}), which also ensures the reproducibility of our approach.

\subsection{Description Prompting at Scale}
\label{sec_prompt}

Recently, many unified learning paradigms simultaneously train detection and grounding data, showing strong transferability to various object-level recognition tasks.
In detail, the previous works, such as GLIP~\cite{GLIP}, GroundingDINO~\cite{GroundingDINO} and UNINEXT~\cite{UNINEXT}, reformulate object detection as phrase grounding by replacing region classifier with word-region alignment.
However, such formulation is difficult to scale up for large-scale detection and grounding with thousand tokens in text prompts.
The main reason is two-fold:
First, the above methods require bidirectional language models, \ie, the pre-trained BERT~\cite{BERT}, to encode text prompts, which can only encode sentences containing at most $512$ tokens.
Second, the above formulation heavily relies on vision-language fusion to perform cross-modality multi-head attention between words and regions at high dimension, \ie, $2,048$, and brings heavy computational costs and GPU memory consumption.
Although the length of text prompt is further limited to $256$ in~\cite{GLIP,GLIPv2,GroundingDINO,UNINEXT}, such fusion module also brings about $0.6 \sim 1.4 \times$ additional memory footprint overall during inference in GLIP~\cite{GLIP}.
A practical solution can split the long text prompts into multiple prompts and query the model multiple times for both training and inference.
However, such a workaround does not address the inherent problem.

To efficiently prompt a large number of vocabularies and sentences all at once, we reverse the widely-used paradigm that cast the classical object detection task into a grounding problem.
Rather, we reformulate visual grounding as object detection to equally unify both localization tasks.
Based on this reformulation, we re-design the text prompt, cross-modality fusion, and vision-language alignment strategy.

\textbf{Independent Prompt.}
Given object classes, such as \textit{Girl} and \textit{Sky}, the previous methods~\cite{GLIP,GLIPv2,GroundingDINO,UNINEXT} concatenate all vocabularies into a single prompt:
``Girl. Sky. \dots'',
in which the corresponding concept embedding is modeled based on its relationship to the other words in the sentence.
The overall prompt embedding $P_\mathrm{voc} \in \Re^{1 \times l \times d}$ has a sequence length of $l$ and an embedding dimension of $d$.
Similarly, the prompts for sentence descriptions are formed as: [``Little child with hat on branch'', ``The big chinchilla'', \dots], thus obtaining embedding $P_\mathrm{sen} \in \Re^{n \times l \times d}$, where $n$ is the sentence number.
We find that the correlation among vocabularies is not necessary, and modeling individual concepts alone is sufficient to identify different instances.

To this end, we blend the individual concepts of vocabularies or sentences as independent text prompts to compute their text embeddings.
Thus, we construct a set of text prompts:
[``Girl'', ``Sky'', ``Little child with hat on branch'', ``The big chinchilla'', \dots],
in which the number of concepts is not limited by the input sequence length of text models.
Those diverse prompts are directly input into directional language models, such as CLIP~\cite{CLIP} and Llama~\cite{Llama2}, getting the prompt embedding $\{P_\mathrm{voc}, P_\mathrm{set} \} \in \Re^{n \times l \times d}$ for $n$ independent text prompts.

\textbf{Sentence-level Embeddings.}
To reduce computational complexity and memory usage, we further compress word-level concept representation to sentence-level prompt embeddings.
Specifically, we aggregate word-level embeddings $\{{P}_\mathrm{voc}, {P}_\mathrm{sen}\}$ to sentence-level ones $\{\bar{P}_\mathrm{voc}, \bar{P}_\mathrm{sen}\} \in \Re^{n \times d}$ via:
$
\bar{P}_{n,d} = \frac{1}{l} \sum_{j=0}^{l}  P_{n,j,d}
$
,
which performs average operator along the length axis.
Albeit word-level prompt embeddings may have more fine-grained information, we find that sentence-level prompt embeddings provide comparable performance, as demonstrated in Sec.~\ref{subsec_Ablations_on_Unified_Detection_and_Grounding}.

\textbf{Gated Cross-modality Interaction.}
The original deep fusion~\cite{GLIP} involves multi-head attention over a large number of input elements for open-vocabulary detection, which usually has thousands of vocabularies to learn.
Thus, we further propose gated cross-modality interaction to restrict different types of prompts from vision-language fusion.
Firstly, the interaction between image features and large-scale vocabularies is prohibitively expensive.
Instead, an all-zero token $\bar{P}_\mathrm{zero} \in \Re^{1 \times 1 \times d}$ serves as a special text embedding and inputs to the fusion module for all given vocabularies.
In this situation, the fusion process is ``static'', as no language information is injected into vision features.
The $\bar{P}_\mathrm{zero}$ could provide explicit instructions to recognize primitive concepts and slightly tune vision feature $\hat{V}_\mathrm{voc}$ and retain original language feature $\bar{P}_\mathrm{voc}$.
For sentence prompts, the corresponding sentence-level embedding $\bar{P}_\mathrm{set}$ are injected into the vision feature $V$, which dynamically updates new vision feature $\hat{V}_\mathrm{set}$ and language feature $\hat{P}_\mathrm{set}$.

The proposed gated fusion enjoys two advantages:
1)
It is feasible to model thousands of detection categories and fuse hundreds of grounding sentences with only a single forward during training and inference.
2)
The previous work has shown that training detection data with a deep fusion module could hurt the zero-shot generalization to novel categories~\cite{GLIP}.
The gated interaction prevents such degeneration by explicitly prohibiting fusion for detection task.

\textbf{Region-sentence Alignment.}
MDETR~\cite{MDETR} first proposes to predict the span of tokens from a text prompt that refers to each matched object, which is so-called word-region alignment.
From a practical perspective, it may not be necessary to detect each word in prompts.
Instead, we predict objects corresponding to the whole prompt, which are a category or sentence.
Concretely,
we compute the alignment scores $S$ between object embeddings $\hat{O}$ and prompt embeddings $\{\bar{P}_\mathrm{voc}, \hat{P}_\mathrm{sen}\}$ as:
$
S = \hat{O} \cdot {(\bar{P}_\mathrm{voc}, \hat{P}_\mathrm{set})}^{\top}
$
,
where $S \in \Re^{n \times m}$.
In such a way, detection categories are fixed anchors in the vision-language common embedding space, as prompt embeddings of vocabularies are not updated in the proposed gated cross-modality interaction.

To compensate for the loss of fine-grained information in sentence-level embedding, we adapt additional irrelevant prompts as negative queries, which imposes the model to have a close ``look'' at target prompts and reject the negative ones. 
In detail, we maintain a history embedding bank and select several embeddings as negative, which are concatenated with positive embeddings for fusion and alignment.

\subsection{Thing-stuff-equalizing Alignment}
\label{sec_thing_stuff}

Recently, MaskFormer~\cite{MaskFormer} and Mask2Former~\cite{Mask2Former} unify thing and stuff segmentation tasks by query-based Transformer architectures and perform mask classification.
However, they are designed for segmentation tasks and usually have unsatisfactory detection performance.
On the other hand, MaskDINO~\cite{MaskDINO} adopts the idea of mask classification from Mask2Former~\cite{Mask2Former} to construct a segmentation branch on DINO~\cite{DINO}.
However, MaskDINO~\cite{MaskDINO} uses different strategies to learn thing and stuff categories, which requires manually identifying vision concepts into two groups ahead.
Thus, it is difficult to incorporate generic semantic-aware segmentation data with large-scale class-agnostic data, such as SA-1B~\cite{SAM}.

To this end, we design a straightforward albeit surprisingly effective solution, where the granularity of the background is equalized to the foreground one, \ie, the model is not aware of the difference between things and stuff.
As stuff categories may mislead the instance-level prediction, stuff regions are composited into multiple disconnected instances, which are treated as standalone samples and aligned with proxy the object-level objective.
During training, we apply connected-component labeling on stuff mask annotations, where subsets of connected components are proxy-ground-truth instances.
For inference, we join all predictions of the same stuff categories as the final results.
Given predicted scores $S \in \Re^{q \times c}$ and masks $M \in \Re^{q \times h \times w}$, the final semantic masks $\hat{M} \in \Re^{c \times h \times w}$ are accumulated as:
$
\hat{M}_{c,h,w} = \sum^{q}_{i=1} S_{i,c} M_{i,h,w}
$
,
where $q$ and $c$ are the number of queries and categories, respectively.

\subsection{Single-Stage Training with Diversity Data}
\label{sec_Aligning}

\textbf{Training Objective.}
To build foundation models that solve a variety of tasks simultaneously, \OurMethod is trained in a single stage without task-specific fine-tuning.
The training objective is a linear combination of classification loss, localization loss, and segmentation loss for the encoder and decoder, respectively:
\begin{equation*}
	\mathcal{L} = \underbrace{\mathcal{L}_\mathrm{class} + \mathcal{L}_\mathrm{bbox} + \mathcal{L}_\mathrm{giou} }_\text{encoder and decoder} + \underbrace{ \mathcal{L}_\mathrm{mask} + \mathcal{L}_\mathrm{dice} }_\text{last layer of decoder}
	,
\end{equation*}
where $\mathcal{L}_\mathrm{class}$ is Focal loss~\cite{FOCALLOSS} to classify foreground and background regions for the encoder and align language and vision embeddings for the decoder.
$\mathcal{L}_\mathrm{bbox}$ and $\mathcal{L}_\mathrm{giou}$ are L1 loss~\cite{FASTERRCNN} and GIoU loss~\cite{GIoU} for box regression, which is applied to both encoder and decoder.
$\mathcal{L}_\mathrm{mask}$ and $\mathcal{L}_\mathrm{dice}$ are cross-entropy loss and dice loss~\cite{V-Net} for mask segmentation, which supervises the last output of decoder only.

\setlength{\textfloatsep}{4pt}
\setlength{\floatsep}{4pt}

\setlength{\abovecaptionskip}{-8pt}
\setlength{\belowcaptionskip}{0pt}

\setlength{\dbltextfloatsep}{4pt}
\setlength{\dblfloatsep}{4pt}

\begin{table*}[t]
\caption{
One suit of weights for open-vocabulary detection on multiple datasets.
``$\varnothing$'' indicates that the task is beyond the model capability.
``--'' indicates that the work does not have a reported number.
}
\footnotesize
\begin{center}
\begin{adjustbox}{max width=1.0\textwidth}
\begin{tabular}{lc|cc|cc|c|c|c|c|cc|cc|cc}
\hline

\multirow{4}{*}{Method}
&\multirow{4}{*}{Backbone}
&\multicolumn{4}{c}{LVIS-1203}
&\multicolumn{1}{|c}{RF100-805}
&\multicolumn{1}{|c}{OID-601}
&\multicolumn{2}{|c}{Objects365-365}
&\multicolumn{4}{|c}{ODinW-314}
&\multicolumn{2}{|c}{COCO-80}
\\

\cline{3-16}

&
&\multicolumn{2}{c}{val}&\multicolumn{2}{|c}{minval}
&\multicolumn{1}{|c}{100 val}
&\multicolumn{1}{|c}{val}
&\multicolumn{1}{|c}{val}&\multicolumn{1}{|c}{minival}
&\multicolumn{2}{|c}{35 val}&\multicolumn{2}{|c}{13 val}
&\multicolumn{2}{|c}{val}
\\

\cline{3-16}

&
&AP$^\mathrm{b}$&AP$^\mathrm{m}$&AP$^\mathrm{b}$&AP$^\mathrm{m}$
&AP$^\mathrm{b}$$_\mathrm{avg}$
&AP$^\mathrm{b}$
&AP$^\mathrm{b}$&AP$^\mathrm{b}$
&AP$^\mathrm{b}$$_\mathrm{avg}$&AP$^\mathrm{b}$$_\mathrm{med}$
&AP$^\mathrm{b}$$_\mathrm{avg}$&AP$^\mathrm{b}$$_\mathrm{med}$
&AP$^\mathrm{b}$&AP$^\mathrm{m}$
\\

\hline

MDETR~\hfilll~\cite{MDETR}&ENB5
&--&$\varnothing$&--&$\varnothing$
&--
&--
&--&--
&10.7&3.0&--&--
&--&$\varnothing$
\\

OWL~\hfilll~\cite{OWLViT}&ViT-L
&34.6&$\varnothing$&--&$\varnothing$
&--
&--
&--&--
&18.8&9.8&--&--
&43.5&$\varnothing$
\\

GLIP~\hfilll~\cite{GLIP}&Swin-L
&26.9&$\varnothing$&37.3&$\varnothing$
&8.6
&61.4
&36.2&39.0
&23.4&11.0&52.1&57.6
&49.8&$\varnothing$
\\

GLIPv2~\hfilll~\cite{GLIPv2}&Swin-H
&--&--&50.1&--
&--
&--
&--&--
&--&--&55.5&--
&\textbf{64.1}&47.4
\\

UNINEXT~\hfilll~\cite{UNINEXT}&ViT-H
&14.0& 12.2&18.3&16.0
&--
&36.1
&23.0&25.5
&--&--&--&--
&60.6&\textbf{51.8}
\\

G-DINO~\hfilll~\cite{GroundingDINO}&Swin-L
&--&$\varnothing$&33.9&$\varnothing$
&--
&--
&--&--
&26.1&18.4&--&--
&60.7&$\varnothing$
\\

OpenSeeD~\hfilll~\cite{OpenSeeD}&Swin-L
&23.0&21.0&--&--
&--
&--
&--&--
&15.2&5.0&--&--
&--&--
\\

\hline

\OurMethod (A)&ViT-L
&55.1&48.7&60.1&53.0
&8.0
&66.5
&46.0&47.5
&25.6&10.0&55.2&64.2
&56.1&47.0
\\

\OurMethod (B)&ViT-L
&57.0&50.5&62.5&55.4
&9.6
&\textbf{68.2}
&47.2&48.9
&\textbf{29.4}&\textbf{16.7}&\textbf{59.8}&\textbf{66.9}
&57.7&48.6
\\

\OurMethod (C)&ViT-L
&56.7&50.7&62.5&55.6
&10.4
&66.6
&46.4&47.9
&29.3&15.4&59.7&66.7
&57.4&48.6
\\

\OurMethod (D)&ViT-L
&\textbf{59.6}&\textbf{53.0}&\textbf{64.7}&\textbf{57.5}
&\textbf{11.9}
&66.7
&\textbf{49.2}&\textbf{51.1}
&28.8&19.9&57.9&64.9
&58.3&49.3
\\

\hline

\end{tabular}
\end{adjustbox}
\end{center}
\label{table_result_ovd}
\end{table*}

\begin{table*}[t]
\caption{
One suit of weights for open-vocabulary segmentation on multiple datasets.
``$\varnothing$'' indicates that the task is beyond the model capability.
``--'' indicates that the work does not have a reported number.
}
\footnotesize
\begin{center}
\begin{adjustbox}{max width=1.0\textwidth}
\begin{tabular}{
lc|c|c|cccc|cc
|c|cc|c|ccc
}

\hline

\multirow{4}{*}{Method}
&\multirow{4}{*}{Backbone}
&\multicolumn{1}{c}{ADE-847}
&\multicolumn{1}{|c}{PC-459}
&\multicolumn{4}{|c}{ADE-150}
&\multicolumn{2}{|c}{SegInW-85}
&\multicolumn{1}{|c}{PC-59}
&\multicolumn{2}{|c}{BDD-40}
&\multicolumn{1}{|c}{VOC-20}
&\multicolumn{3}{|c}{Cityscapes-19}
\\

\cline
{3-17}

&
&\multicolumn{1}{c}{val}
&\multicolumn{1}{|c}{val}
&\multicolumn{4}{|c}{val}
&\multicolumn{2}{|c}{25 val}
&\multicolumn{1}{|c}{val}
&\multicolumn{2}{|c}{val}
&\multicolumn{1}{|c}{val}
&\multicolumn{3}{|c}{val}
\\

\cline
{3-17}

&
&mIoU
&mIoU
&PQ&AP$^\mathrm{m}$&AP$^\mathrm{b}$&mIoU
&AP$^\mathrm{m}_\mathrm{avg}$&AP$^\mathrm{m}_\mathrm{med}$
&mIoU
&PQ&mIoU
&mIoU
&PQ&AP$^\mathrm{m}$&mIoU
\\

\hline

X-Decoder~\hfilll~\cite{X-Decoder}&DaViT-L
&9.2
&16.1
&21.8&13.1&--&{29.6}
&22.3&32.3
&\textbf{64.0}
&17.8&47.2
&\textbf{97.7}
&38.1&24.9&\textbf{52.0}\\

OpenSeeD~\hfilll~\cite{OpenSeeD}&Swin-L
&--
&--
&19.7&15.0&17.7&23.4
&36.1&38.7
&--
&\textbf{19.4}&\textbf{47.4}
&--
&\textbf{41.4}&\textbf{33.2}&47.8\\

OpenSeg~\hfilll~\cite{OpenSeg}&ENB7

&8.8
&12.2
&--&--&--&28.6
&--&--
&48.2
&--&--
&72.2
&--&--&--
\\

OVSeg~\hfilll~\cite{OVSeg}&Swin-B
&9.0
&12.4
&-&--&--&29.6
&--&--
&55.7
&--&--
&94.5
&--&--&--
\\

\hline

\OurMethod (A)&ViT-L
&4.1
&15.9
&26.6&23.8&28.7&28.9
&47.4&48.0
&51.2
&15.2&41.8
&90.4
&28.6&27.4&37.9
\\

\OurMethod (B)&ViT-L
&9.2
&21.0
&26.4&23.5&28.5&29.0
&46.4&53.7
&58.3
&13.4&35.3
&95.8
&26.9&26.6&37.2
\\

\OurMethod (C)&ViT-L
&\textbf{9.4}
&20.1
&26.1&23.8&28.8&28.5
&47.8&49.9
&58.6
&16.2&43.9
&95.5
&32.8&30.7&42.6
\\

\OurMethod (D)&ViT-L
&9.2
&\textbf{21.8}
&\textbf{27.2}&\textbf{24.4}&\textbf{29.6}&\textbf{30.0}
&\textbf{49.6}&\textbf{52.2}
&58.5
&17.4&45.7
&96.5
&33.3&30.3&44.2
\\

\hline

\end{tabular}
\end{adjustbox}
\end{center}
\label{table_result_ovs}
\end{table*}

\begin{table*}[t]
\caption{
One suit of weights for visual grounding on D$^3$.
$\varnothing$ indicates that the task is beyond the model capability.
``--'' indicates that the work does not have a reported number.
}
\footnotesize
\begin{center}
\begin{adjustbox}{max width=1.0\textwidth}

\begin{tabular}{lc | cc | cc | cc | cc | cc | cc | cc | cc | c c}
\hline

\multirow{4}{*}{Method} & \multirow{4}{*}{Backbone} & \multicolumn{6}{c|}{Intra-scenario} & \multicolumn{12}{c}{Inter-scenario} \\

\cline{3-20}

&&
\multicolumn{2}{c}{\textsl{Full}} & \multicolumn{2}{c}{\textsl{Presence}} & \multicolumn{2}{c|}{\textsl{Absence}} & \multicolumn{2}{c}{\textsl{Full}} & \multicolumn{2}{c}{\textsl{Presence}} & \multicolumn{2}{c}{\textsl{Absence}} & \multicolumn{2}{|c}{\textsl{Full}} & \multicolumn{2}{c}{\textsl{Presence}} & \multicolumn{2}{c}{\textsl{Absence}} \\

\cline{3-20}

&&
AP$^\mathrm{b}$&AP$^\mathrm{m}$&AP$^\mathrm{b}$&AP$^\mathrm{m}$&AP$^\mathrm{b}$&AP$^\mathrm{m}$&AP$^\mathrm{b}$&AP$^\mathrm{m}$&AP$^\mathrm{b}$&AP$^\mathrm{m}$&AP$^\mathrm{b}$&AP$^\mathrm{m}$&AR$^\mathrm{b}$&AR$^\mathrm{m}$&AR$^\mathrm{b}$&AR$^\mathrm{m}$&AR$^\mathrm{b}$&AR$^\mathrm{m}$\\

\hline

OFA~\hfilll~\cite{OFA} & R152 & 4.2 &$\varnothing$& 4.1 &$\varnothing$& 4.6 &$\varnothing$& 0.1 &$\varnothing$& 0.1 &$\varnothing$& 0.1 &$\varnothing$ & 17.1 &$\varnothing$& 16.7 &$\varnothing$& 18.4 &$\varnothing$\\

CORA~\hfilll~\cite{CORA} & R50 & 6.2 &$\varnothing$& 6.7 &$\varnothing$& 5.0 &$\varnothing$& 2.0 &$\varnothing$& 2.2 &$\varnothing$& 1.3 &$\varnothing$ & 10.0 &$\varnothing$& 10.5 &$\varnothing$& 8.7 &$\varnothing$\\
OWL-ViT~\hfilll~\cite{OWLViT} & ViT-L & 9.6 &$\varnothing$& 10.7 &$\varnothing$& 6.4 &$\varnothing$& 2.5 &$\varnothing$& 2.9 &$\varnothing$& 2.1 &$\varnothing$ & 17.5 &$\varnothing$& 19.4 &$\varnothing$& 11.8 &$\varnothing$\\

UNINEXT~\hfilll~\cite{UNINEXT} & ViT-H & 20.0 &--& 20.6 &--& 18.1 &--& 3.3 &--& 3.9 &--& 1.6 &-- & 45.3 &--& 46.7 &--& 41.4 &--\\
G-DINO~\hfilll~\cite{GroundingDINO} & Swin-B & 20.7 &$\varnothing$& 20.1 &$\varnothing$& {22.5} &$\varnothing$& 2.7 &$\varnothing$& 2.4 &$\varnothing$& 3.5 &$\varnothing$ & 51.1 &$\varnothing$& 51.8 &$\varnothing$& 48.9  &$\varnothing$\\

OFA-DOD~\hfilll~\cite{DOD} & R101 & {21.6} &$\varnothing$& {23.7} &$\varnothing$& 15.4 &$\varnothing$& {5.7} &$\varnothing$& {6.9} &$\varnothing$& 2.3 &$\varnothing$ & 47.4 &$\varnothing$& 49.5 &$\varnothing$& 41.2 &$\varnothing$\\

\hline

\OurMethod (A)&ViT-L
&25.1&22.6&24.5&22.0&26.9&24.4
&16.4&14.8&15.9&14.3&17.9&16.3&63.1&57.4&63.5&57.5&62.1&57.1
\\
\OurMethod (B)&ViT-L
&30.0&26.8&29.9&26.6&30.3&27.3
&20.0&18.0&20.5&18.4&\textbf{18.6}&\textbf{16.8}&79.3&70.8&79.0&70.2&79.9&72.4
\\
\OurMethod (C)&ViT-L
&27.8&25.6&27.9&25.5&27.3&25.8
&20.4&\textbf{18.7}&21.2&\textbf{19.3}&18.1&16.9&79.9&72.8&80.1&72.6&79.0&73.3
\\
\OurMethod (D)&ViT-L
&\textbf{37.5}&\textbf{34.4}&\textbf{38.8}&\textbf{35.4}&\textbf{33.9}&\textbf{31.7}
&\textbf{21.0}&18.3&\textbf{22.0}&\textbf{19.3}&17.9&15.4&\textbf{82.7}&\textbf{73.7}&\textbf{82.8}&\textbf{73.5}&\textbf{82.4}&\textbf{74.3}\\

\hline
\end{tabular}

\end{adjustbox}
\end{center}
\label{table_result_vg}
\end{table*}

\begin{table*}[t]
\caption{
Ablation study of unified detection and grounding.
O-W: object-word fusion.
R-S: region-sentence fusion.
}
\footnotesize
\begin{center}
\begin{adjustbox}{max width=1.0\textwidth}
\begin{tabular}{cccc|cc|cc|cc|cc|cc|cc|cc|cc|cc|cc|cc}
\hline

\multirow{4}{*}{Training}
&\multirow{4}{*}{Fusion}
&\multirow{4}{*}{Bank}
&\multirow{4}{*}{Text}
&\multicolumn{2}{c}{COCO}
&\multicolumn{2}{|c}{LVIS}
&\multicolumn{6}{|c}{RefCOCO}
&\multicolumn{6}{|c}{RefCOCO+}
&\multicolumn{6}{|c}{RefCOCOg}\\

\cline{5-26}

&&&
&\multicolumn{2}{c}{val}
&\multicolumn{2}{|c}{val}
&\multicolumn{2}{|c}{val}		
&\multicolumn{2}{|c}{testA}		
&\multicolumn{2}{|c}{testB}
&\multicolumn{2}{|c}{val}		
&\multicolumn{2}{|c}{testA}		
&\multicolumn{2}{|c}{testB}	
&\multicolumn{2}{|c}{umd-val}	
&\multicolumn{2}{|c}{umd-test}		
&\multicolumn{2}{|c}{google-val}
\\

\cline{5-26}

&&&
&AP$^\mathrm{b}$&AP$^\mathrm{m}$
&AP$^\mathrm{b}$&AP$^\mathrm{m}$
&P@1&oIoU&P@1&oIoU&P@1&oIoU
&P@1&oIoU&P@1&oIoU&P@1&oIoU
&P@1&oIoU&P@1&oIoU&P@1&oIoU
\\

\hline

\multirow{4}{*}{RefC}
&\crossmarknew&\crossmarknew&CLIP
&--&--&--&--&80.5&65.1&84.6&69.0&72.5&58.5&70.4&53.7&77.6&59.4&59.1&44.2&79.0&62.6&72.8&55.2&72.2&55.7\\ 

&\crossmarknew&\crossmarknew&T5
&--&--&--&--&74.2&57.5&78.5&61.7&69.3&53.2&57.7&41.6&64.0&46.4&49.8&35.1&67.3&49.6&61.0&42.6&61.8&44.2\\

&O-W&\crossmarknew&CLIP
&--&--&--&--&85.5&71.7&89.1&74.1&81.3&67.1&73.4&57.7&80.7&63.4&64.4&48.9&83.0&67.9&78.0&61.7&78.0&61.9\\

&O-W&\crossmarknew&BERT
&--&--&--&--&84.6&70.7&88.6&73.6&79.6&66.4&72.4&57.1&79.0&61.7&62.2&47.2&82.9&67.4&77.7&60.9&78.0&62.1\\

\hline

\multirow{3}{*}{RefC}
&R-S&\crossmarknew&CLIP
&--&--&--&--&80.7&65.8&85.0&69.7&74.8&60.3&68.9&52.3&76.2&58.7&59.2&43.5&74.1&56.7&72.5&55.2&70.7&53.8\\

&R-S&\checkmarknew&CLIP
&--&--&--&--&82.7&68.2&87.1&72.2&77.4&62.9&72.3&56.7&79.7&62.6&61.3&46.1&79.5&63.5&74.8&57.6&73.8&57.5\\

&R-S&\checkmarknew&Llama2
&--&--&--&--&85.3&71.5&89.1&75.3&80.0&66.4&73.9&58.9&81.8&66.4&64.0&49.0&85.1&68.8&74.5&57.5&75.2&59.9\\

\hline

\multirow{5}{*}{+COCO}
&\crossmarknew&\crossmarknew&CLIP&
50.9&43.7&--&--&81.6&67.0&85.5&71.2&77.9&63.4&69.8&53.7&76.3&58.8&61.4&45.3&78.9&62.0&74.9&57.1&74.7&57.1\\

&\crossmarknew&\crossmarknew&T5&
50.9&43.7&--&--&78.3&62.6&81.7&65.3&75.8&60.2&60.6&45.0&66.3&49.6&53.6&38.6&71.9&54.5&67.7&48.6&68.9&50.2\\

&O-W&\crossmarknew&CLIP&
49.2&42.2&--&--&85.1&71.5&87.9&73.4&83.4&69.8&71.8&56.0&77.5&60.5&64.5&49.2&80.3&64.7&76.8&60.3&77.5&61.1\\

&O-W&\crossmarknew&BERT&
50.2&43.2&--&--&85.6&71.5&87.8&73.3&83.2&68.8&71.2&55.4&77.0&60.2&63.8&48.8&81.1&66.2&77.7&60.9&79.1&62.9\\

&R-S&\crossmarknew&CLIP&
50.9&43.7&--&--&84.0&70.7&88.0&74.4&80.7&67.7&73.2&57.8&79.6&63.5&64.5&49.5&80.7&64.7&76.2&59.0&77.0&60.3\\

\hline

\multirow{5}{*}{+LVIS}
&\crossmarknew&\crossmarknew&CLIP
&--&--&36.8&32.9&78.6&63.7&83.2&68.8&73.4&59.0&69.5&53.2&75.3&57.7&61.0&45.2&77.6&61.1&73.6&55.1&73.9&56.7\\

&O-W&\crossmarknew&CLIP
&--&--&33.0&29.6&86.6&73.5&88.6&74.9&83.5&70.9&74.7&59.5&79.8&63.3&66.3&50.9&82.9&68.2&79.4&62.9&79.9&63.7\\

&O-W&\crossmarknew&BERT
&--&--&25.6&23.0&84.8&70.9&87.0&72.9&82.3&68.2&71.7&56.1&76.6&59.9&64.7&49.5&79.9&64.3&77.6&60.8&77.2&61.4\\

&R-S&\crossmarknew&CLIP
&--&--&37.8&34.1&83.3&69.7&87.2&73.8&78.2&64.8&73.3&57.9&80.2&63.7&64.5&48.7&79.4&63.6&75.6&58.0&75.6&58.9\\

&R-S&\checkmarknew&CLIP
&--&--&37.8&33.9&85.2&71.8&88.5&74.8&81.6&68.2&75.0&59.9&81.1&64.4&66.1&50.3&81.2&66.0&77.6&59.7&78.0&62.1\\

\hline
\end{tabular}
\end{adjustbox}
\end{center}
\label{table_ablation_fusion_align}
\end{table*}

\begin{table*}[t]
\caption{
Comparison of computational cost in terms of the proportions of decreased speed~(FPS) and increased memory~(GB) for for cross-modality interaction.
``$\varnothing$'' indicates that the task is beyond the model capability.
}
\footnotesize
\begin{center}
\begin{adjustbox}{max width=0.95\textwidth}
\begin{tabular}{c|cc|cc|cc|cc|cc|cc}
\hline

\multirow{6}{*}{Model}
&\multicolumn{10}{c|}{Inference}
&\multicolumn{2}{c}{\multirow{4}{*}{Train}}
\\

\cline{2-11}

&\multicolumn{4}{c|}{Detection}
&\multicolumn{6}{c|}{Grounding}
&&
\\

\cline{2-11}

&\multicolumn{2}{c}{COCO 80 classes}
&\multicolumn{2}{|c}{LVIS 1203 classes}
&\multicolumn{2}{|c}{1 sentences}
&\multicolumn{2}{|c}{128 sentences}
&\multicolumn{2}{|c|}{1280 sentences}
&&
\\

\cline{2-13}

&FPS&GB
&FPS&GB
&FPS&GB
&FPS&GB
&FPS&GB
&FPS&GB
\\

\hline

\multirow{2}{*}{GLIP-T}
&$\downarrow 47\%$
&$\uparrow 140\%$
&$\downarrow 98\%$
&$\uparrow 140\%$
&$\downarrow 47\%$
&$\uparrow 140\%$
&\multirow{2}{*}{$\varnothing$}&\multirow{2}{*}{$\varnothing$}&\multirow{2}{*}{$\varnothing$}&\multirow{2}{*}{$\varnothing$}
&$\downarrow 41\%$
&$\uparrow 39\%$
\\

&\color{gray} 4.8 $\veryshortarrow$ 2.5
&\color{gray} 1.0 $\veryshortarrow$ 2.4
&\color{gray} 4.4 $\veryshortarrow$ 0.08
&\color{gray} 1.0 $\veryshortarrow$ 2.4
&\color{gray} 4.8 $\veryshortarrow$ 2.5
&\color{gray} 1.0 $\veryshortarrow$ 2.4
&&&&
&\color{gray} 2.7 $\veryshortarrow$ 1.6
&\color{gray} 11.5 $\veryshortarrow$ 16.0
\\

\hline

\multirow{2}{*}{\OurMethod-R50}
&$\downarrow 32\%$
&$\uparrow 61\%$
&$\downarrow 34\%$
&$\uparrow 48\%$

&$\downarrow 29\%$
&$\uparrow 61\%$
&$\downarrow 39\%$
&$\uparrow 80\%$
&$\downarrow 76\%$
&$\uparrow 270\%$

&$\downarrow 35\%$
&$\uparrow 25\%$
\\

&\color{gray} 10.3 $\veryshortarrow$ 6.9
&\color{gray} 1.3 $\veryshortarrow$ 2.2
&\color{gray} 10.1 $\veryshortarrow$ 6.6
&\color{gray} 1.7 $\veryshortarrow$ 2.5

&\color{gray} 9.3 $\veryshortarrow$ 6.5
&\color{gray} 1.3 $\veryshortarrow$ 2.1
&\color{gray} 9.2 $\veryshortarrow$ 5.5
&\color{gray} 1.3 $\veryshortarrow$ 2.4
&\color{gray} 9.1 $\veryshortarrow$ 2.1
&\color{gray} 1.3 $\veryshortarrow$ 5.1

&\color{gray} 1.1 $\veryshortarrow$ 0.7
&\color{gray} 7.7 $\veryshortarrow$ 9.7
\\

\hline

\multirow{2}{*}{GLIP-L}
&$\downarrow 40\%$
&$\uparrow 60\%$
&$\downarrow 96\%$
&$\uparrow 60\%$
&$\downarrow 40\%$
&$\uparrow 60\%$
&\multirow{2}{*}{$\varnothing$}&\multirow{2}{*}{$\varnothing$}&\multirow{2}{*}{$\varnothing$}&\multirow{2}{*}{$\varnothing$}
&$\downarrow 30\%$
&$\uparrow 18\%$
\\

&\color{gray} 0.5 $\veryshortarrow$ 0.3
&\color{gray} 4.8 $\veryshortarrow$ 7.7
&\color{gray} 0.5 $\veryshortarrow$ 0.017
&\color{gray} 4.8 $\veryshortarrow$ 7.7
&\color{gray} 0.5 $\veryshortarrow$ 0.3
&\color{gray} 4.8 $\veryshortarrow$ 7.7
&&&&
&\color{gray} 1.1 $\veryshortarrow$ 0.8
&\color{gray} 19.7 $\veryshortarrow$ 23.4
\\

\hline

\multirow{2}{*}{\OurMethod-L}
&$\downarrow 11\%$
&$\uparrow 19\%$
&$\downarrow 8\%$
&$\uparrow 19\%$

&$\downarrow 12\%$
&$\uparrow 19\%$
&$\downarrow 14\%$
&$\uparrow 24\%$
&$\downarrow 46\%$
&$\uparrow 89\%$

&$\downarrow 15\%$
&$\uparrow 12\%$
\\

&\color{gray} 2.2 $\veryshortarrow$ 2.0
&\color{gray} 4.2 $\veryshortarrow$ 5.0
&\color{gray} 2.1 $\veryshortarrow$ 1.9
&\color{gray} 4.4 $\veryshortarrow$ 5.2

&\color{gray} 2.1 $\veryshortarrow$ 1.8
&\color{gray} 4.2 $\veryshortarrow$ 5.0
&\color{gray} 2.1 $\veryshortarrow$ 1.8
&\color{gray} 4.2 $\veryshortarrow$ 5.3
&\color{gray} 2.0 $\veryshortarrow$ 1.1
&\color{gray} 4.3 $\veryshortarrow$ 8.1

&\color{gray} 0.6 $\veryshortarrow$ 0.5
&\color{gray} 10.9 $\veryshortarrow$ 12.3
\\

\hline
\end{tabular}
\end{adjustbox}
\end{center}
\label{table_ablation_computational_cost}
\end{table*}

\begin{table*}[t]
\caption{
Ablations study of thing-stuff-equalizing learning with various training data.
}
\footnotesize
\begin{center}
\begin{adjustbox}{max width=0.92\textwidth}
\begin{tabular}{ccc|c|cccccccccc}
\hline

\multirow{2}{*}{Training Data}
&\multirow{2}{*}{Equalize Thing\&Stuff}
&\multirow{2}{*}{Step}
&\multicolumn{1}{c}{COCO-Stuff}
&\multicolumn{10}{|c}{COCO}
\\

\cline{4-14}

&&
&mIoU
&mIoU
&PQ&SQ&RQ&PQ$^\mathrm{th}$&SQ$^\mathrm{th}$&RQ$^\mathrm{th}$&PQ$^\mathrm{st}$&SQ$^\mathrm{st}$&RQ$^\mathrm{st}$
\\

\hline

\multirow{2}{*}{COCO-Panoptic}
&\crossmarknew&90k
&33.9
&55.7
&47.4&81.0&57.2&53.8&82.9&64.3&37.8&78.1&46.5
\\

&\checkmarknew&90k
&37.6
&58.0
&48.2&81.3&57.9&54.4&83.5&64.6&38.8&78.1&47.7
\\

\hline

\multirow{2}{*}{COCO-Instance}
&\crossmarknew&90k
&43.0
&52.2
&44.8&80.9&54.1&52.3&83.0&62.5&33.5&77.6&41.4
\\

&\checkmarknew&90k
&45.5
&54.5
&45.5&81.3&54.7&52.3&83.4&62.2&35.3&78.0&43.3
\\

\hline

\multirow{2}{*}{LVIS, COCO-Stuff}
&\crossmarknew&375k
&42.0
&52.8
&44.8&80.9&54.1&52.3&83.0&62.5&33.5&77.6&41.4
\\

&\checkmarknew&375k
&46.2
&55.4
&45.5&81.2&54.6&52.3&83.4&62.1&35.2&78.0&43.3
\\

\hline

\multirow{2}{*}{LVIS, COCO-Stuff, RefC}
&\crossmarknew&375k
&42.8
&53.8&44.8&81.7&53.9&49.3&82.9&58.8&38.0&80.0&46.5
\\

&\checkmarknew&375k
&46.1
&55.2&45.3&81.9&54.4&49.2&83.0&58.5&39.3&80.3&48.1
\\

\hline
\end{tabular}
\end{adjustbox}
\end{center}
\label{table_ablation_thing_stuff}
\end{table*}

\begin{table}[t]
\caption{
Ablation study of using SA-1B~\cite{SAM}.
}
\footnotesize
\begin{center}
\begin{adjustbox}{max width=0.99\textwidth}
\begin{tabular}{cc|cc|cc}
\hline

\multirow{2}{*}{Training Data}
&\multirow{2}{*}{Step}
&\multicolumn{2}{c}{COCO val}
&\multicolumn{2}{|c}{LVIS val}
\\

\cline{3-6}

&
&AP$^\mathrm{b}$&AP$^\mathrm{m}$
&AP$^\mathrm{b}$&AP$^\mathrm{m}$
\\

\hline

COCO&90k&
50.0&42.6&--&--\\

COCO, SA-1B&180k&
50.1&43.4&10.3&9.2\\

\hline

LVIS&180k&
--&--&37.3&33.4\\

LVIS, SA-1B&375k&
--&--&39.1&35.3\\

\hline

\end{tabular}
\end{adjustbox}
\end{center}
\label{table_ablation_sa1b}
\end{table}

\begin{table}[t]
\caption{
System comparisons of instance segmentation.
}
\footnotesize
\begin{center}
\begin{adjustbox}{max width=0.45\textwidth}
\begin{tabular}{l|c|c|cc|cc}
\hline

\multirow{2}{*}{Method}
&\multirow{2}{*}{Backbone}
&\multirow{2}{*}{Size}
&\multicolumn{2}{c}{COCO val}
&\multicolumn{2}{|c}{LVIS val}
\\

\cline{4-7}

&&
&AP$^\mathrm{b}$&AP$^\mathrm{m}$
&AP$^\mathrm{b}$&AP$^\mathrm{m}$
\\

\hline

MaskDINO~\hfilll~\cite{MaskDINO}&SwinL&$1024$&59.0&52.3&--&--\\

ViTDet~\hfilll~\cite{ViTDet}&ViT-H&$1024$&60.4&52.0&53.4&48.1\\

EVA-02~\hfilll~\cite{EVA-02}&ViT-L&$1536$&62.3&53.8&60.1&53.5\\

\OurMethod&ViT-L&$1536$&\textbf{62.7}&\textbf{54.1}&\textbf{60.9}&\textbf{55.3}\\

\hline
\end{tabular}
\end{adjustbox}
\end{center}
\label{table_single_dataset}
\end{table}

\textbf{Training Data.}
With the above loss function, $10$ datasets with different annotation types are employed to train \OurMethod.
For object detection, \OurMethod simultaneously learns common vocabularies from MS COCO~\cite{MSCOCO}, Objects365~\cite{Objects365} and OpenImages~\cite{OpenImages}, and long-tailed LVIS~\cite{LVIS}.
OpenImages and LVIS are also federated datasets with sparse annotations.
For image segmentation, apart from mask annotations in MS COCO and LVIS, \OurMethod also learns class-agnostic segmentation data from SA-1B~\cite{SAM}, which contains both things and stuff without semantic labels.
For visual grounding, we joint Visual Genome~\cite{VisualGenome}, RefCOCO/+/g~\cite{REFCOCO,REFCOCOg}, GQA~\cite{GQA}, Flickr30K~\cite{FlickrEntities}, and PhraseCut~\cite{PhraseCut}.

To handle the diverse data and meet the requirement of single-stage training, we propose three principles for multi-dataset and multi-task learning:
First, the well-annotated detection and segmentation data supervise all classification losses, localization losses, and even segmentation losses when there exist pixel-level annotations.
For federated datasets, such as LVIS and OpenImages, a federated loss is integrated into classification losses $\mathcal{L}_\mathrm{class}$ in the decoder.
For class-agnostic data from SA-1B, the classification losses $\mathcal{L}_\mathrm{class}$ in the decoder is not trained.
Second, grounding data is only used to learn classification losses $\mathcal{L}_\mathrm{class}$ in the decoder, as most grounding data does not exhaustively annotate all images with all objects and the bounding boxes are often not as accurate as detection data.
For grounding data with segmentation annotations, such as RefCOCO/+/g, all loss functions in the decoder are trained.
Third, we set the dataset sampling ratio to $1.0$ if the dataset has more than $100$K images and $0.1$ otherwise.
We list the configures of sampling ratios and loss weights for all datasets in Tab.~\ref{table_config_data} of the supplement material.

\textbf{Image-centri Grounding Samples.}
The previous methods construct grounding samples of the form $\{I, T, B\}$, where $I$ is an image, $T$ is a phrase that describes an instance in $I$, and $B$ is the corresponding bounding box.
However, compared to detection training where all annotations in an image are trained all at once, the above region-centri format makes grounding training inefficient, as the model is supervised by a single instance for each sample.
\OurMethod is feasible to handle multiple phrase prompts in a single forward pass during the training and inference.
Thus, we gather the grounding samples in the image-centri form of $\{I, (T_i, B_i), \dots, (T_n, B_n) \}$, which groups the grounding annotations.
The new image-centri format significantly reduces the number of training iterations while the model still receives the same amount of supervision. 
For example, there are on average $92$ box-level annotations~(region descriptions and object instances) per image in Visual Genome.
The proposed image-centri format leads to $92\times$ speedup over the traditional region-centri format.
During training, to prevent multiple phrases referring to the same object, we apply NMS to all boxes with random scores.

\section{Experiments}

We show that \OurMethod serves as a strong general-purpose vision system after training, \ie, one model weight for all.
Specifically, \OurMethod is directly evaluated on over $160$ datasets to detect, segment, and ground without fine-tuning.
Implementation details are in Sec.~\ref{sec_implemnetation_details} of the supplement material.

\textbf{\OurMethod (A)} is built on DETA~\cite{DETA} with our designs replacing the corresponding modules.
It is based on the ViT-L~\cite{ViT} and only trained on detection and segmentation data, including COCO, LVIS, Objects365, OpenImages, and Visual Genome.
\textbf{\OurMethod (B)} is enhanced with Visual Genome region descriptions and RefCOCO/+/g.
It is designed to verify the effectiveness of grounding data.
\textbf{\OurMethod (C)} adds class-agnostic data SA-1B for training.
\textbf{\OurMethod (D)} further include incorporates GQA, PhraseCut, and Flickr30k.
Details on datasets are in Sec.~\ref{sec_data_details} of the supplement material.
All \OurMethod models are jointly trained on the corresponding datasets in a single stage without any fine-tuning.
We list data usages in Tab.~\ref{table_info_data} of supplementary.
Compared to other methods, our framework feeds the least number of images to models during training.

\subsection{One Model Weight for All}
To investigate the generalization ability of \OurMethod, we evaluate our models on various domain- and task-specific datasets.

\textit{Object Detection.}
In Tab.~\ref{table_result_ovd}, we evaluate on the well-established benchmarks, including large-vocabulary and long-tailed dataset, \eg, LVIS~\cite{LVIS}, and common object detection datasets, \eg, Objects365~\cite{Objects365}, OpenImages~\cite{OpenImages} and MSCOCO~\cite{MSCOCO}.
\OurMethod achieves the state-of-the-art or competitive performance across all benchmarks simultaneously.
We find that the existing methods, such as GLIP, OWL, and UNINEXT, while including Objecst365 during training, fall short in delivering strong performance on Objecst365.
The proposed \OurMethod is notably superior to them, proving that \OurMethod remembers all seen concepts well.

In addition, we further introduce Roboflow~\cite{Roboflow100} and ODinW~\cite{GLIP}, which consist of $100$ and $35$ datasets, respectively, with different imagery domains, to evaluate generalizability under real-world scenarios.
\OurMethod achieves a new SotA on both Roboflow and ODinW, validating \OurMethod can handle a large-scale of diverse concepts in the wild.

\textit{Image Segmentation.}
We then compare \OurMethod with the previous works on various segmentation tasks.
We report PQ, AP$^\mathrm{m}$, and mIoU for panoptic, instance, and semantic segmentation, respectively.
Overall, \OurMethod achieves significantly better performance on PC-459, ADE20K, and SegInW with $459$, $150$, and $85$ categories, respectively, and comparable performance for BDD, VOC, and Cityscapes with only $40$, $20$, and $19$ categories, respectively.
SegInW consists of $25$ diverse segmentation datasets.
The results demonstrate that \OurMethod has superior generalization ability to detect and segment a wide range of object categories in real-world scenarios.
Note that we only use instance-level annotations for training, which puts \OurMethod at a disadvantage in the evaluation of panoptic-level results in terms of PQ.
This is because the panoptic task requires non-overlap instance predictions, while \OurMethod produces overlapping segments.

\textit{Visual Grounding.}
We evaluate the model’s ability to ground objects in natural language on description detection dataset~(D$^3$)~\cite{DOD}.
Following work in~\cite{DOD}, we evaluate each image with only the descriptions that existed in the image as intra-scenario.
For inter-scenario, all references in the dataset are used to query the models.
It is noted that, for both intra-scenario and inter-scenario setting, other methods only processes a single description for one forward, while \OurMethod only needs a single forward to query all references.

As demonstrated in Tab.~\ref{table_result_vg}, \OurMethod outperforms all existing methods for all metrics in D$^3$.
Specifically, \OurMethod achieves significant improvement in the inter-scenario evaluation.
\OurMethod is naturally capable of rejecting irrelevant prompts, while the previous methods almost completely fail in this setting.
This demonstrated that formulating visual grounding as open-vocabulary detection with sentence-object vision-language fusion and alignment not only significantly improves the efficiency of prompting a lot of text queries, but also obtains a strong ability to reject negative references.
And \textsl{Full}, \textsl{Presence}, and \textsl{Absence} denote evaluation on all descriptions, presence descriptions only, and absence descriptions only.
We find that \OurMethod is less biased toward the presence descriptions and well handles the absence descriptions.
We further conduct experiments on RefCOCO/+/g in Tab.~\ref{table_result_vg2} of supplement material, \OurMethod surpasses all other methods without further fine-tuning.

\subsection{Ablations on Unified Detection and Grounding}
\label{subsec_Ablations_on_Unified_Detection_and_Grounding}
Different cross-modality interactions could have a large impact on description references, and vocabulary concepts usually robust to the fusion and alignment strategies.
Thus, we first perform an in-depth study of various module combinations on both object detection and visual grounding.
Unless otherwise specified, we conduct ablation experiments with R-50~\cite{ResNet} with an input size of $800 \times 1,333$.

\textit{Region-sentence formulation \vs Object-word formulation.}
In the first part of Tab.~\ref{table_ablation_fusion_align}, we find that the fusion module is helpful for visual grounding task and boost the performance for all metrics.
While sentence-level text embedding may lose fine-grained information, region-sentence fusion still achieves comparable performance with object-word fusion, as shown in the second part of Tab.~\ref{table_ablation_fusion_align}.

\textit{Effectiveness of History Embedding Bank.}
To compensate for the loss of fine-grained information in sentence-level text embedding, we add the proposed history embedding bank to region-sentence fusion, and the performance of the visual grounding task is significantly improved.
This is because the text embedding bank introduces negative descriptions that do not exist in the images, and imposes the model to learn relevant information from language guidance.

\textit{Effectiveness of Joint Detection and Grounding Training.}
We further combine RefCoco/+/g~{RefC} with MSCOCO or LVIS with a dataset ratio of $1 : 1$.
In the third and fourth parts of Tab.~\ref{table_ablation_fusion_align}, the proposed region-sentence fused models with text embedding bank catch up with, and even surpass, the counterparts of object-word fusion.
We also find that, for large-vocabulary LVIS, object-word fusion could deteriorate the performance of object detection, while the proposed formulation also improves the detection results, demonstrating the effectiveness of \OurMethod.

\textit{Effectiveness of Text Models.}
We also ablate the influence of text models in Tab.~\ref{table_ablation_fusion_align}.
We find that contrastive pre-trained models, \ie, CLIP~\cite{CLIP}, have better performance than BERT~\cite{BERT} and T5~\cite{Raffel2020} with masked image modeling for both detection and grounding tasks.
Meanwhile, large language models~\cite{Llama2} also improve grounding results.

\textit{Computational Cost.}
We test the additional computational cost of the cross-modality fusion by measuring decreased speed and increased memory in percentage.
Following GLIP~\cite{GLIP}, for training, we use $2$ and $1$ images per batch for \OurMethod-R50 and \OurMethod-ViT-L, respectively.
For inference, we use batch size $1$ for all models.
For a fair comparison, we disable the segmentation part and use $5$ feature maps with strides ranging from $8$ to $128$.
Tab.~\ref{table_ablation_computational_cost} shows that \OurMethod brings less proportion of additional computational costs than GLIP.
It is feasible to query \OurMethod with a large number of prompts all at once, which validates the efficiency of the proposed gated cross-modality interaction.

\subsection{Ablations on Thing-stuff-equalizing Alignment}
We further conduct ablations on the proposed thing-stuff-equalizing alignment, which formulates both thing and stuff categories as instance-level learning.
We train our models on COCO panoptic segmentation and evaluate them on COCO stuff and panoptic segmentation.
In Tab.~\ref{table_ablation_thing_stuff}, the result indicates that our unification significantly enhances both segmentation performances in terms of PQ and mIoU.
We also combine LVIS and COCO stuff as training data, and the results show that our improvement is also helpful for large-scale vocabularies.

To validate the compatibility of class-agnostic and semantic-aware annotations, we jointly train with SA-1B, LVIS, and MSCOCO.
As shown in Tab.~\ref{table_ablation_sa1b}, the result indicates that SA-1B significantly helps the instance-level detection and segmentation for large-scale vocabularies.

\subsection{Performance on Single Dataset}
We further evaluate the performance of \OurMethod on a single benchmark \textbf{without} additional training data.
For the long-tailed LVIS, we choose FedLoss~\cite{Zhou2021b} as the classification loss to remedy the impact of unbalanced data distribution.
We compare the performances of instance segmentation in Tab.~\ref{table_single_dataset}, and all methods are \textbf{not} pre-trained on Objects365.
Our method suppresses all other models and achieves state-of-the-art results on MSCOCO and LVIS.

\section{Conclusion}

We present a \OurMethodFullName to perform diverse tasks, \ie, detection, segmentation, and grounding, as an instance-level sentence-object matching paradigm.
\OurMethod is trained on broad data with natural and challenging characteristics, such as Zipfian distribution of categories, federated annotations, anything segmentation, and mixed vocabulary and sentence concepts.
The extensive experiments show that \OurMethod outperforms (or is on par with) the existing SotA models with only one-suit of weights, proving that an effective yet universal perception for anything prompting and alignment at scale is indeed feasible.

{
    \small
    \bibliographystyle{ieeenat_fullname}
    \bibliography{main/library_format.bib}
}

\clearpage
\setcounter{page}{1}
\maketitlesupplementary

\section{Appendix}

\subsection{Model Structure Details}
\label{sec_sturcture_details}
In this section, we compare \OurMethod with other models from the perspective of model structure.
As shown in Tab.~\ref{table_info_framework}, our model has a significantly different framework.
Compared to GLIPv2~\cite{GLIPv2} and UNINEXT~\cite{UNINEXT}, \OurMethod uses a smaller input size for the long side and has only half the number of parameters.

\begin{table*}[t]
\caption{
The relevant information of different models including the backbone, base detector, text encoder, and image size.
}
\footnotesize
\begin{center}
\begin{adjustbox}{max width=0.88\textwidth}
\begin{tabular}{l|c|c|c|c|c}
\toprule

\multirow{2}{*}{Method}&\multirow{2}{*}{Backbone}&\multirow{2}{*}{Base Model}&\multirow{2}{*}{Text Encoder}&\multicolumn{2}{|c}{Image Size}\\

\cmidrule{5-6}
&&&&Short&Long\\

\midrule
{MDETR~\hfilll~\cite{MDETR}}							&ENB5 \hfilll (30M)		&DETR			&RoBERTa&480 $\sim$ 800	&1333\\

\midrule
{GLIP~\hfilll~\cite{GLIP}}								&Swin-L \hfilll (197M)	&DyHead			&BERT	&480 $\sim$ 800	&1333\\

\midrule
{GLIPv2~\hfilll~\cite{GLIPv2}}							&CoSwin-H \hfilll (637M)	&DyHead			&CLIP	&480 $\sim$ 800	&1333\\

\midrule
{UNINEXT~\hfilll~\cite{UNINEXT}}						&ViT-H \hfilll (632M)	&DINO			&BERT	&320 $\sim$ 800	&1333\\

\midrule
{G-DINO~\hfilll~\cite{GroundingDINO}}					&Swin-L \hfilll (197M)	&DINO			&BERT	&480 $\sim$ 800	&1333\\

\midrule
{X-Decoder~\hfilll~\cite{X-Decoder}}					&DaViT-L\hfilll (196M) 	&Mask2Former	&UniCL	&224, 1024 &224, 1024\\

\midrule
{OpenSeeD~\hfilll~\cite{OpenSeeD}}						&Swin-L \hfilll (197M)	&MaskDINO		&UniCL	&1024			&1024\\

\midrule
{SEEM~\hfilll~\cite{SEEM}}								&DaViT-L \hfilll (196M) &X-Decoder		&UniCL	&800			&1333\\

\midrule
{HIPIE~\hfilll~\cite{HIPIE}} 							&ViT-H \hfilll (637M)	&UNINEXT		&BERT	&800 $\sim$ 1024&1333\\

\midrule
{ODISE~\hfilll~\cite{ODISE}}							&UNet \hfilll (860M)	&Mask2Former 	&CLIP	&1024			&1024\\

\midrule
{\OurMethod (A)}												&ViT-L \hfilll (307M)	&DETA			&CLIP	&1024			&1024\\

{\OurMethod (B)}												&ViT-L \hfilll (307M)	&DETA			&CLIP	&1024			&1024\\

{\OurMethod (C)}												&ViT-L \hfilll (307M)	&DETA			&CLIP	&1024			&1024\\

{\OurMethod (D)}												&ViT-L \hfilll (307M)	&DETA			&CLIP	&1024			&1024\\

\bottomrule
\end{tabular}
\end{adjustbox}
\end{center}
\label{table_info_framework}
\end{table*}

\subsection{Training Data Details}
\label{sec_data_details}

We compare the data usage of various \OurMethods and other models in Tab.~\ref{table_info_data}.
It shows that our method consumes the least images during training while achieving superior performance.
The main reason is two-fold:
First, we enable to query \OurMethods with a large number of prompts, which speeds up model coverage.
Second, we design image-centri format to group grounding data, efficiently reducing training iterations and speedup coverage.
Based on the three principles in Sec.~\ref{sec_Aligning}, we configure the sampling ratios and loss weights for all datasets as shown in Tab.~\ref{table_config_data}.

\begin{table*}[ht]
\caption{
A detailed list of training data for different models.
O365: Objects365.
OID: OpenImages Detection.
VG: Visual Genome.
INB: ImageNetBoxes.
RefC: RefCOCO/+/g.
}
\footnotesize
\begin{center}
\begin{adjustbox}{max width=0.99\textwidth}
\begin{tabular}{l|c|c|c|c|c}
\toprule

\multirow{2}{*}{Method}&\multirow{2}{*}{Stage}&\multicolumn{2}{c|}{Train Data (Group by annotation types)}&\multirow{2}{*}{Batch Size}&Image Consumption\\

\cmidrule{3-4}
&&Instance-level&Image-level&&\#Epoch $\times$ \#Image or Batch Size $\times$ \#Iteration\\

\midrule
{MDETR~\hfilll~\cite{MDETR}}
&\rom{1}
&COCO, RefC, VG, GQA, Flickr30k
&--
&64
&52M  ( 40 Ep $\times$ 1.3M Img )
\\

\midrule
{GLIP~\hfilll~\cite{GLIP}}
&\rom{1}
&O365, OID, VG, INB, COCO, RefC, VG, GQA, Flickr30k
&Cap24M
&64
&64M   ( 64 Bs $\times$ 1M Iter )
\\

\midrule
\multirow{2}{*}{
{GLIPv2}
}
~\hfilll			
\multirow{2}{*}{
{~\cite{GLIPv2}}
}			
&\rom{1}
&O365, OID, VG, INB, COCO, RefC, VG, GQA, Flickr30k
&\multirow{2}{*}{Cap16M}
&64
&64M   ( 64 Bs $\times$ 1M Iter )
\\
&\rom{2}
&COCO, LVIS, PhraseCut&
&64
&5.36M   ( 24 Ep $\times$ 0.2M Img + 8 Ep $\times$0.07M Img )
\\

\midrule
\multirow{3}{*}
{
{UNINEXT}
}
~\hfilll
\multirow{3}{*}
{
{~\cite{UNINEXT}}
}
&\rom{1}&Objects365
&\multirow{3}{*}{--}
&64&21.8M   ( 64 Bs $\times$ 340741 Iter )
\\
&\rom{2}
&COCO, RefC
&
&32
&2.9M   ( 32 Bs $\times$ 91990 Iter )
\\
&\rom{3}
&COCO, RefC, SOT\&VOS, MOT\&VIS, R-VOS
&
&32
&5.7M   ( 32 Bs $\times$ 180000 Iter )
\\

\midrule
{GroundingDINO\hfilll~\cite{GroundingDINO}}
&\rom{1}
&COCO, O365, OID, RefC, Flickr30k, VG
&Cap4M
&64
&--		\\

\midrule
{X-Decoder~\hfilll~\cite{X-Decoder}}
&\rom{1}&COCO, RefC
&Cap4M
&32, 1024
&200M   ( 50 Ep $\times$ 4M Img ) \\

\midrule
{OpenSeeD~\hfilll~\cite{OpenSeeD}}
&\rom{1}
&COCO, O365
&--
&32, 64
&48M  ( 30 Ep $\times$ 1.8M Img )	\\

\midrule
{SEEM~\hfilll~\cite{SEEM}}
&\rom{1}
&COCO, LVIS, RefC
&--&--&--
\\

\midrule
{\OurMethod (A)}								&\rom{1}&COCO, LVIS, O365, OID, VG						&--&16	&11.52M   ( 16 Bs $\times$ 0.72M Iter )	
\\

{\OurMethod (B)}								&\rom{1}&COCO, LVIS, O365, OID, VG, RefC			&--&16	&17.28M   ( 16 Bs $\times$ 1.08M Iter )
\\

{\OurMethod (C)}								&\rom{1}&COCO, LVIS, O365, OID, VG, RefC, SA-1B	&--&16	&17.28M   ( 16 Bs $\times$ 1.08M Iter )
\\

{\OurMethod (D)}								&\rom{1}&COCO, LVIS, O365, OID, VG, RefC, SA-1B, GQA, PhraseCut, Flickr30k	&--&64	&17.28M   ( 64 Bs $\times$ 0.27M Iter )
\\

\bottomrule
\end{tabular}
\end{adjustbox}
\end{center}
\label{table_info_data}
\end{table*}

\begin{table*}[t]
\caption{
Training data configures.
SR denotes the sampling ratio, and FL denotes federated loss.
}
\footnotesize
\begin{center}
\begin{adjustbox}{max width=0.99\textwidth}
\begin{tabular}{ccc|ccc|cccccccc}
\toprule

\multirow{3}{*}{Dataset}&\multirow{3}{*}{SR}&\multirow{3}{*}{FL}&\multicolumn{8}{c}{Loss Weights}\\

\cmidrule{4-11}

&&&\multicolumn{3}{c}{Encoder}&\multicolumn{5}{|c}{Decoder}\\
\cmidrule{4-11}
&&&$\mathcal{L}_\mathrm{class}$&$\mathcal{L}_\mathrm{bbox}$&$\mathcal{L}_\mathrm{giou}$&$\mathcal{L}_\mathrm{class}$&$\mathcal{L}_\mathrm{bbox}$&$\mathcal{L}_\mathrm{giou}$&$\mathcal{L}_\mathrm{mask}$&$\mathcal{L}_\mathrm{dice}$\\
\midrule

LVIS	&1.0&\checkmark&1&5&2&1&5&2&5&5\\

COCO Instance	&1.0&&1&5&2&1&5&2&5&5\\

COCO Stuff	&1.0&&1&5&2&1&5&2&5&5\\

Objects365			&1.0&&1&5&2&1&5&2&5&5\\

OpenImages		&1.0&\checkmark&1&5&2&1&5&2&5&5\\

Visual Genome		&1.0&&0&0&0&1&0&0&0&0\\

SA-1B				&1.0&&1&5&2&0&5&2&5&5\\

RefCOCO/+/g				&0.1&&0&5&2&1&5&2&5&5\\

GQA					&0.1&&0&0&0&1&0&0&0&0\\

Flickr30K			&0.1&&0&0&0&1&0&0&0&0\\

PhraseCut			&0.1&&0&0&0&1&0&0&0&0\\

\bottomrule
\end{tabular}
\end{adjustbox}
\end{center}
\label{table_config_data}
\end{table*}

\subsection{Implementation Details}
\label{sec_implemnetation_details}

We build on DETA~\cite{DETA} to implement our model.
DETA has a simpler alternative training mechanism to learn an easier decoding function with IoU-based label assignment.
We use $900$ queries and $6$ encoder and decoder layers.
For the visual backbone, we adopt pre-trained ViT-L~\cite{EVA-02} by default and also use ReseNet-50~\cite{ResNet} in our ablation studies.
We adopt the pre-trained large model in EVA-CLIP~\cite{EVA-CLIP} for the language backbone.
We use the AdamW~\cite{AdamW} optimizer with a weight decay of $0.05$ and a learning rate $2e-4$, which is decayed at $0.88$ fractions of the total number of steps by $10$.
We also compare our structure to other models for the largest model size in Tab.~\ref{table_info_framework}.

For data augmentation, we use the default large-scale jittering~\cite{CopyPaste} augmentation with a random scale sampled from the range $0.1$ to $2.0$ for all datasets.
For COCO~\cite{MSCOCO}, instead of panoptic mask annotations, we utilize $80$-category instance-level and $53$-category semantic-level annotations as the supervision signal.
We also apply repeat factor sampling~\cite{LVIS} and copy-paste augmentation~\cite{CopyPaste} on LVIS~\cite{LVIS}.
Detailed descriptions of implementation are available in the supplementary material.

\subsection{Additional Result of Visual Grounding}

We further conduct experiments on RefCOCO/+/g datasets with other models that only require a single stage of training.
As shown in Tab.~\ref{table_result_vg2}, \OurMethod surpasses all other methods with large performance gaps.

\begin{table*}[t]
\caption{
One suit of weights for visual grounding on RefCOCO/+/g.
``$\varnothing$'' indicates that the task is beyond the model capability.
``--'' indicates that the work does not have a reported number.
}
\footnotesize
\begin{center}
\begin{adjustbox}{max width=0.99\textwidth}
\begin{tabular}{lc|cc|cc|cc|cc|cc|cc|cc|cc|cc}
\toprule

\multirow{4}{*}{Method}
&\multirow{4}{*}{Backbone}
&\multicolumn{6}{c}{RefCOCO}		
&\multicolumn{6}{|c}{RefCOCO+}
&\multicolumn{6}{|c}{RefCOCO}
\\

\cmidrule{3-20}

&
&\multicolumn{2}{c}{val}		&\multicolumn{2}{|c}{testA}		&\multicolumn{2}{|c}{testB}
&\multicolumn{2}{|c}{val}		&\multicolumn{2}{|c}{testA}		&\multicolumn{2}{|c}{testB}
&\multicolumn{2}{|c}{umd-val}	&\multicolumn{2}{|c}{umd-test}	&\multicolumn{2}{|c}{google-val}
\\

\cmidrule{3-20}

&
&P@1&oIoU&P@1&oIoU&P@1&oIoU
&P@1&oIoU&P@1&oIoU&P@1&oIoU
&P@1&oIoU&P@1&oIoU&P@1&oIoU
\\

\midrule

MDETR~\hfilll~\cite{MDETR}&ENB5
&73.4&$\varnothing$&--&$\varnothing$&--&$\varnothing$&58.8&$\varnothing$&--&$\varnothing$&--&$\varnothing$&57.1&$\varnothing$&--&$\varnothing$&--&$\varnothing$\\
GLIP~\hfilll~\cite{GLIP}&Swin-T
&50.4&$\varnothing$&54.3&$\varnothing$&43.8&$\varnothing$&49.5&$\varnothing$&52.7&$\varnothing$&44.5&$\varnothing$&66.0&$\varnothing$&66.8&$\varnothing$&--&$\varnothing$\\

GroundingDINO~\hfilll~\cite{GroundingDINO}&Swin-T
&73.9&$\varnothing$&74.8&$\varnothing$&59.2&$\varnothing$&66.8&$\varnothing$&69.9&$\varnothing$&56.0&$\varnothing$&71.0&$\varnothing$&72.0&$\varnothing$&--&$\varnothing$\\

KOSMOS-2~\hfilll~\cite{KOSMOS2}&ViT-L
&52.3&$\varnothing$&57.4&$\varnothing$&47.2&$\varnothing$&45.4&$\varnothing$&50.7&$\varnothing$&42.2&$\varnothing$&60.5&$\varnothing$&61.6&$\varnothing$&--&$\varnothing$\\

\midrule
\OurMethod (A)&ViT-L
&34.2&25.1&34.8&28.0&36.1&25.7&33.5&26.3&32.3&26.6&36.0&26.0&38.9&28.1&40.5&28.3&39.4&28.4
\\
\OurMethod (B)&ViT-L
&83.3&70.2&88.4&76.0&77.7&63.9&74.0&59.4&82.0&67.6&62.9&47.8&79.9&62.8&79.9&62.8&\textbf{80.5}&\textbf{64.3}
\\
\OurMethod (C)&ViT-L
&79.8&66.3&86.8&74.0&76.2&61.8&72.2&56.6&78.4&64.1&60.9&45.6&79.8&63.2&79.5&61.2&79.5&62.6
\\
\OurMethod (D)&ViT-L
&\textbf{84.6}&\textbf{72.3}&\textbf{89.2}&\textbf{77.7}&\textbf{80.9}&\textbf{68.4}&\textbf{76.4}&\textbf{61.9}&\textbf{82.4}&\textbf{68.0}&\textbf{66.5}&\textbf{51.2}&\textbf{80.0}&\textbf{64.2}&\textbf{80.1}&\textbf{63.2}&79.9&63.3
\\

\bottomrule
\end{tabular}
\end{adjustbox}
\end{center}
\label{table_result_vg2}
\end{table*}

\subsection{Visualization}
In this subsection, we demonstrate the generalization ability to
various datasets and flexibility to support task compositions for \OurMethod with qualitative visualizations.

\begin{figure*}[t!]
	\centering
	
	\begin{subfigure}[t]{0.3\textwidth}
		\centering
		\includegraphics[height=1.2in]{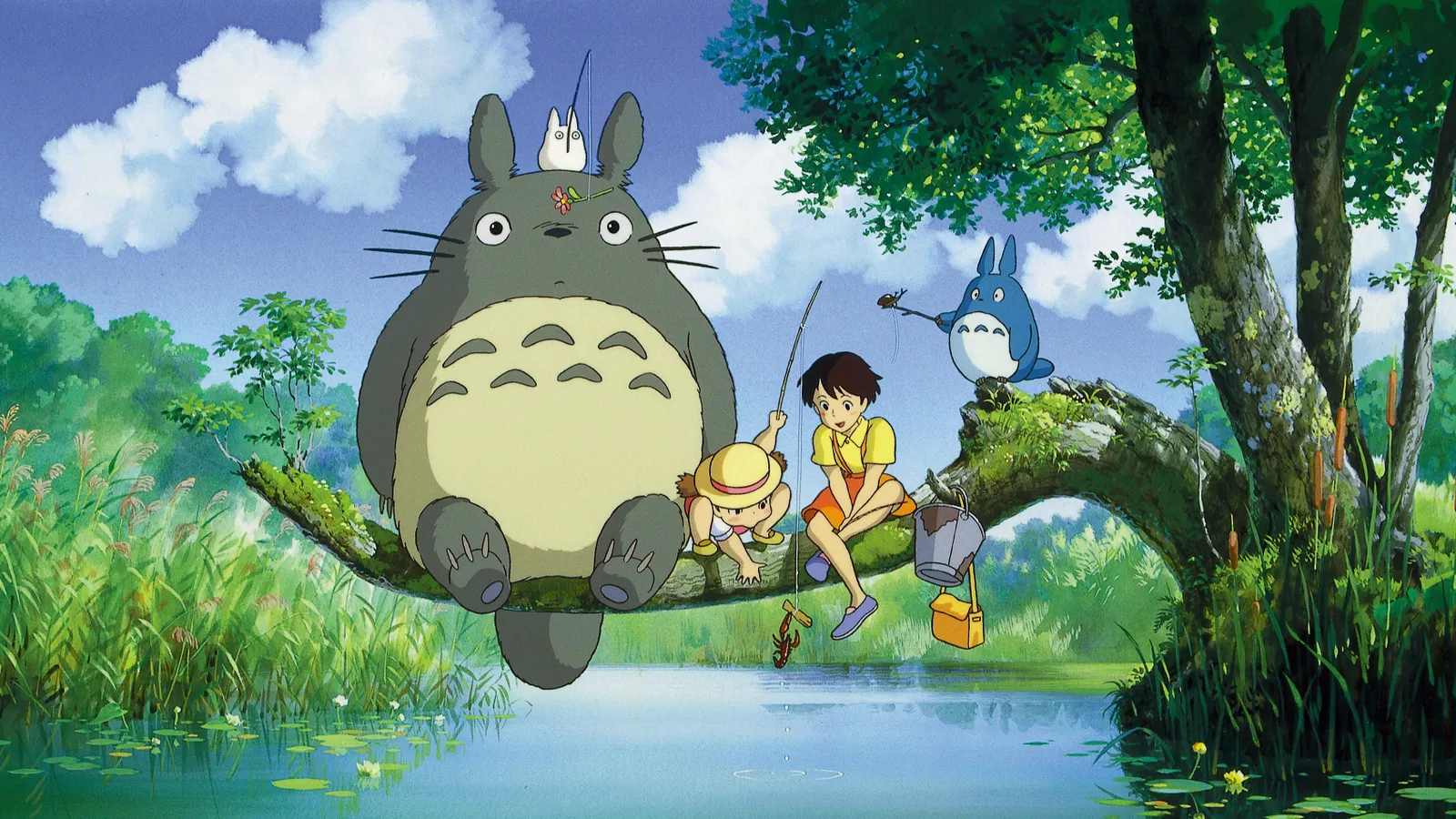}
		\caption{Original Image.}
	\end{subfigure}%
	~ 
	\begin{subfigure}[t]{0.3\textwidth}
		\centering
		\includegraphics[height=1.2in]{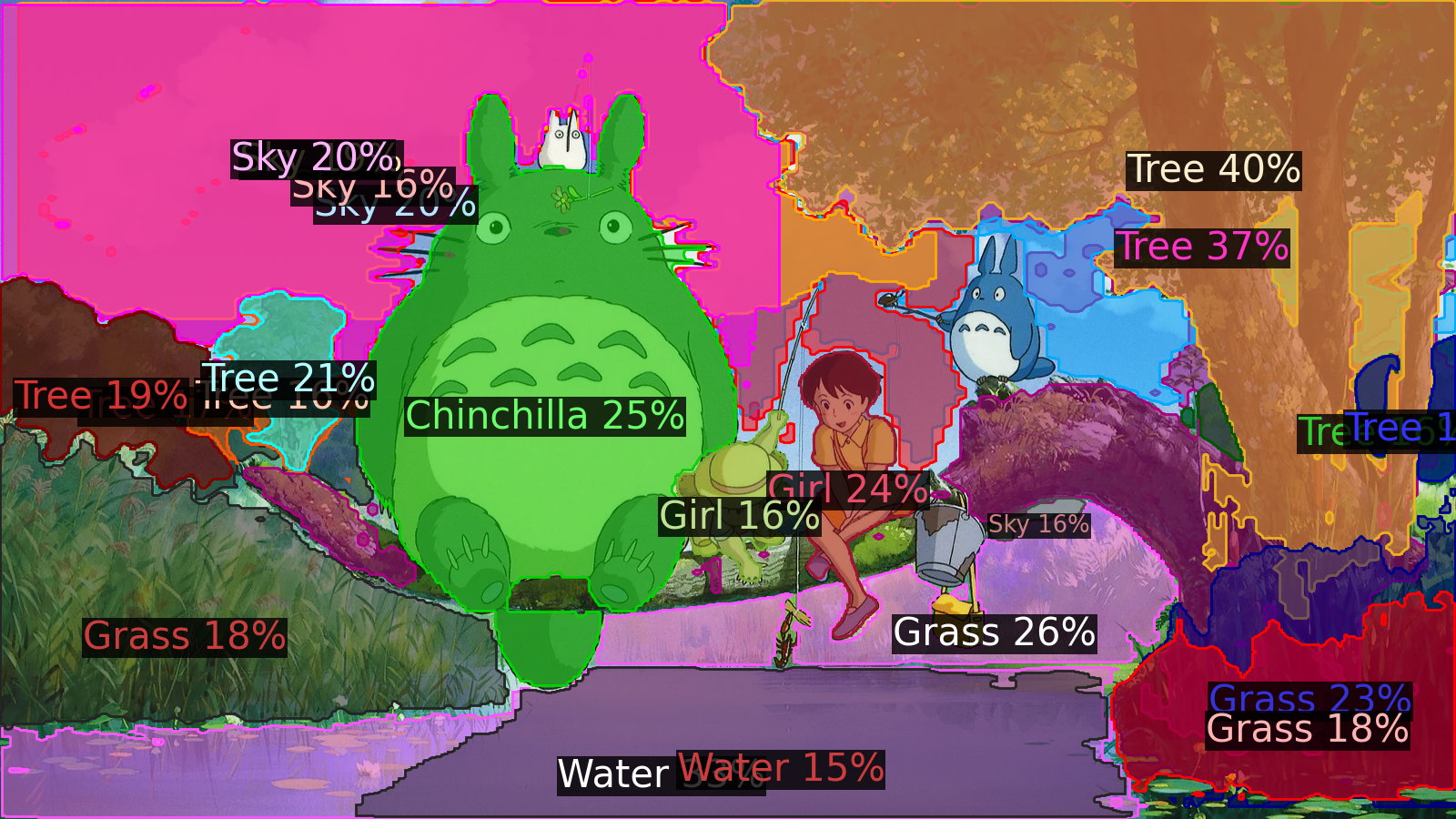}
		\caption{Instance Segmentation.}
	\end{subfigure}%
	~ 
	\begin{subfigure}[t]{0.3\textwidth}
		\centering
		\includegraphics[height=1.2in]{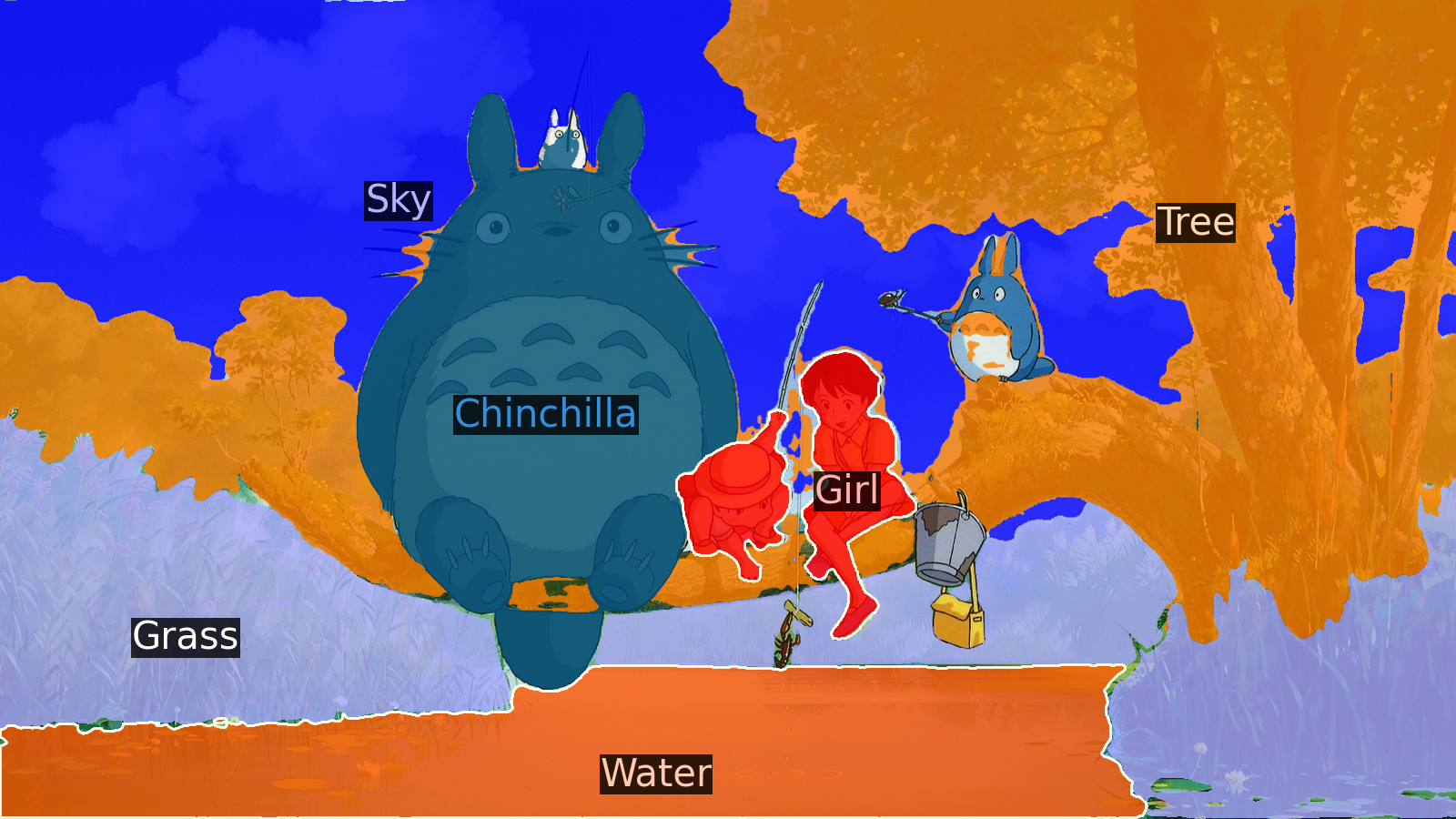}
		\caption{Semantic Segmentation.}
	\end{subfigure}

	\vspace{10pt}
	\caption{
		Visualizations of model outputs for instance and semantic segmentation tasks.
		All results are inferred in a single forward with prompts of \{``Sky'', ``Water'', ``Tree'', ``Chinchilla'', ``Grass'', ``Girl''\}.
	}
	\label{fig_vis_thing_stuff}
\end{figure*}

In Fig.~\ref{fig_vis_thing_stuff}, we first visualize the model outputs for instance and semantic segmentation tasks.
Noted that all results for both tasks are the same outputs from \OurMethod-D, except for different post-processing.
For instance segmentation, we apply non-maximum suppression on predicted regions.
For semantic segmentation, we further accumulate the semantic masks for the same concepts as described in subsec.~\ref{sec_thing_stuff}.

\begin{figure*}[t!]
	\centering
	
	\begin{subfigure}[t]{0.4\textwidth}
		\centering
		\includegraphics[height=1.2in]{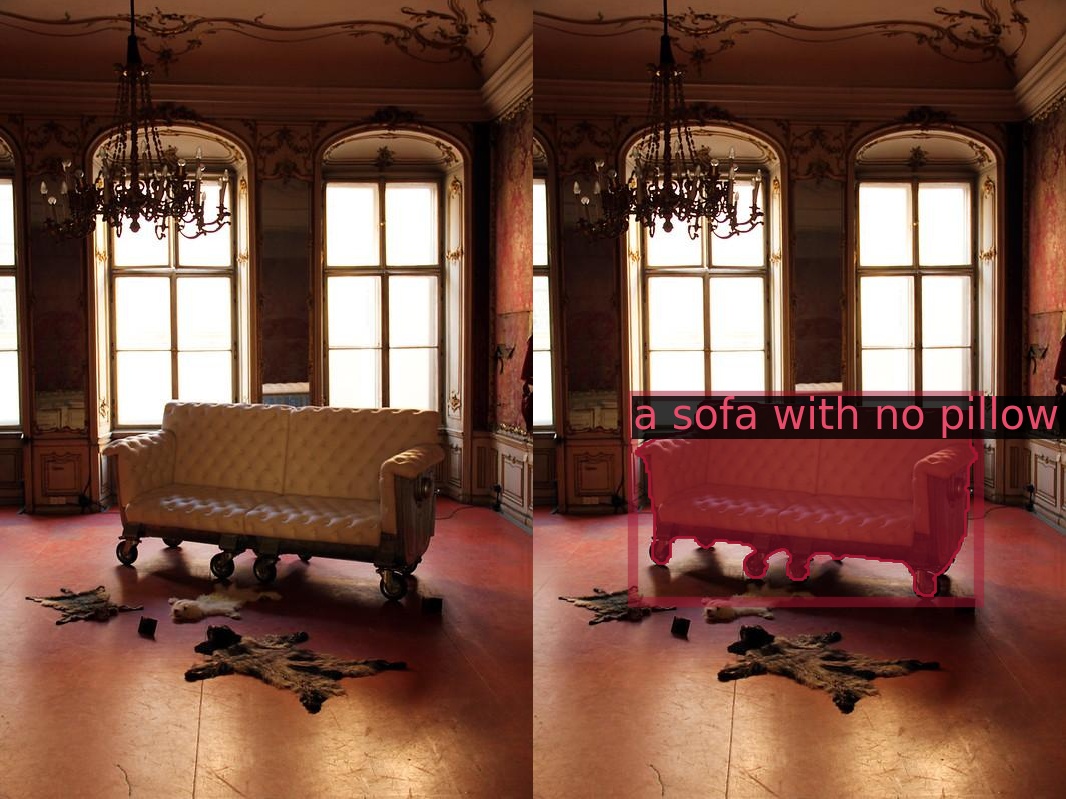}
		\caption{``a sofa with no pillow on it in the room''}
	\end{subfigure}%
	~ 
	\begin{subfigure}[t]{0.6\textwidth}
		\centering
		\includegraphics[height=1.2in]{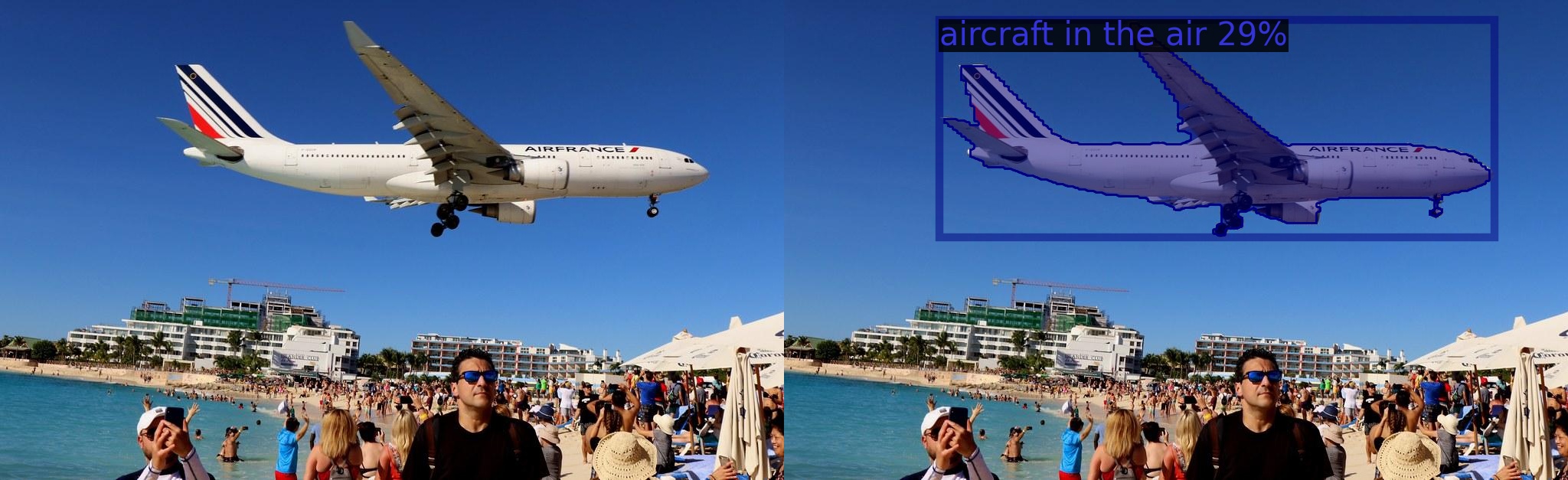}
		\caption{``aircraft in the air''}
	\end{subfigure}
	
	\begin{subfigure}[t]{0.4\textwidth}
		\centering
		\includegraphics[height=1.2in]{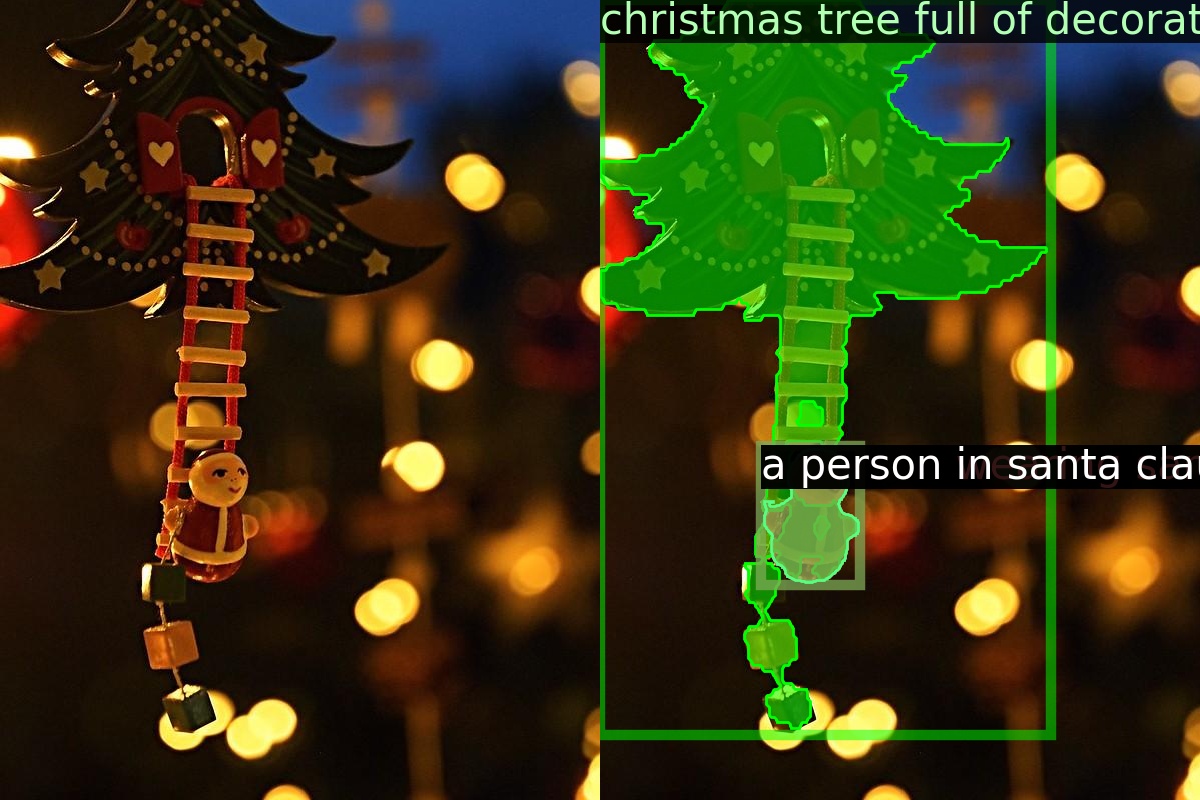}
		\caption{``christmas tree full of decorations'', ``a person in santa claus clothes without bags''}
	\end{subfigure}%
	~
	\begin{subfigure}[t]{0.6\textwidth}
		\centering
		\includegraphics[height=1.2in]{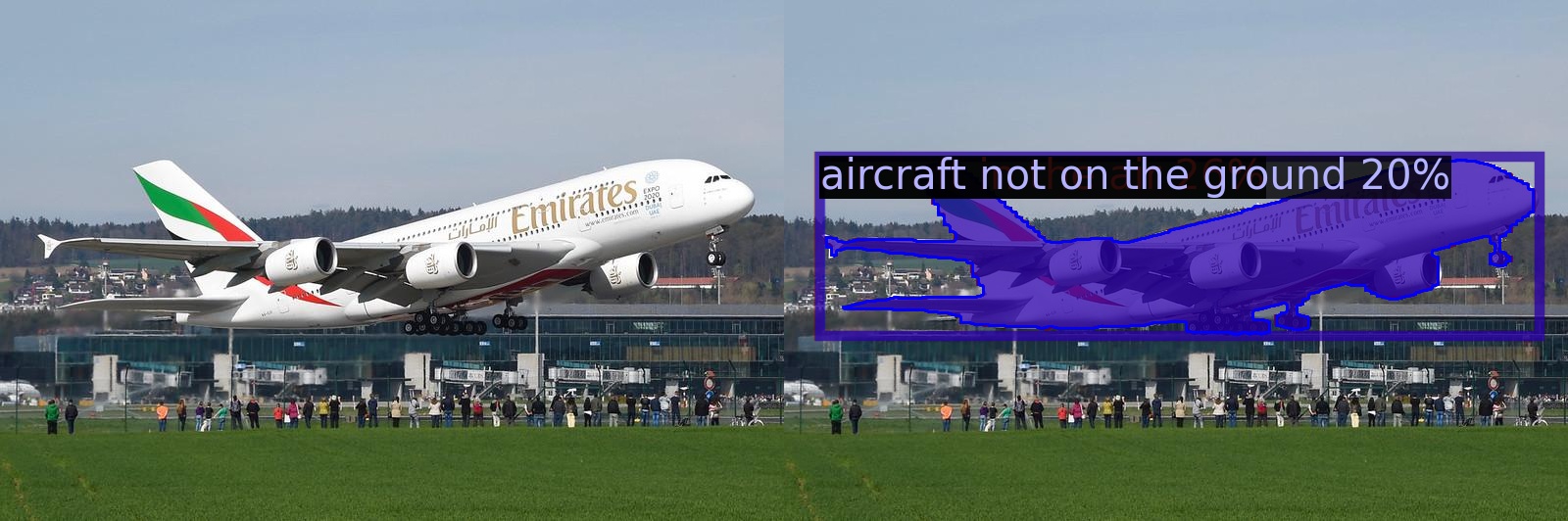}
		\caption{``aircraft not on the ground''}
	\end{subfigure}
	
	\begin{subfigure}[t]{0.4\textwidth}
		\centering
		\includegraphics[height=1.2in]{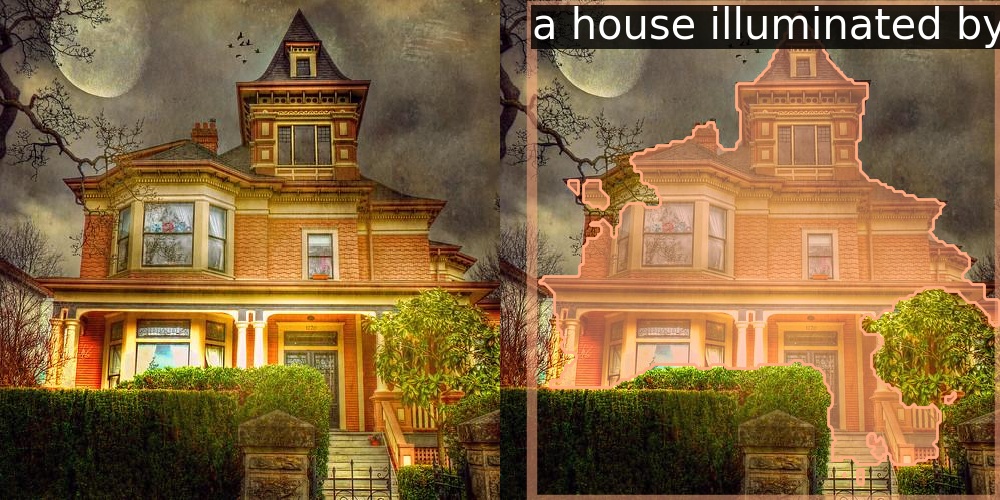}
		\caption{``a house illuminated by the moon''}
	\end{subfigure}%
	~ 
	\begin{subfigure}[t]{0.6\textwidth}
		\centering
		\includegraphics[height=1.2in]{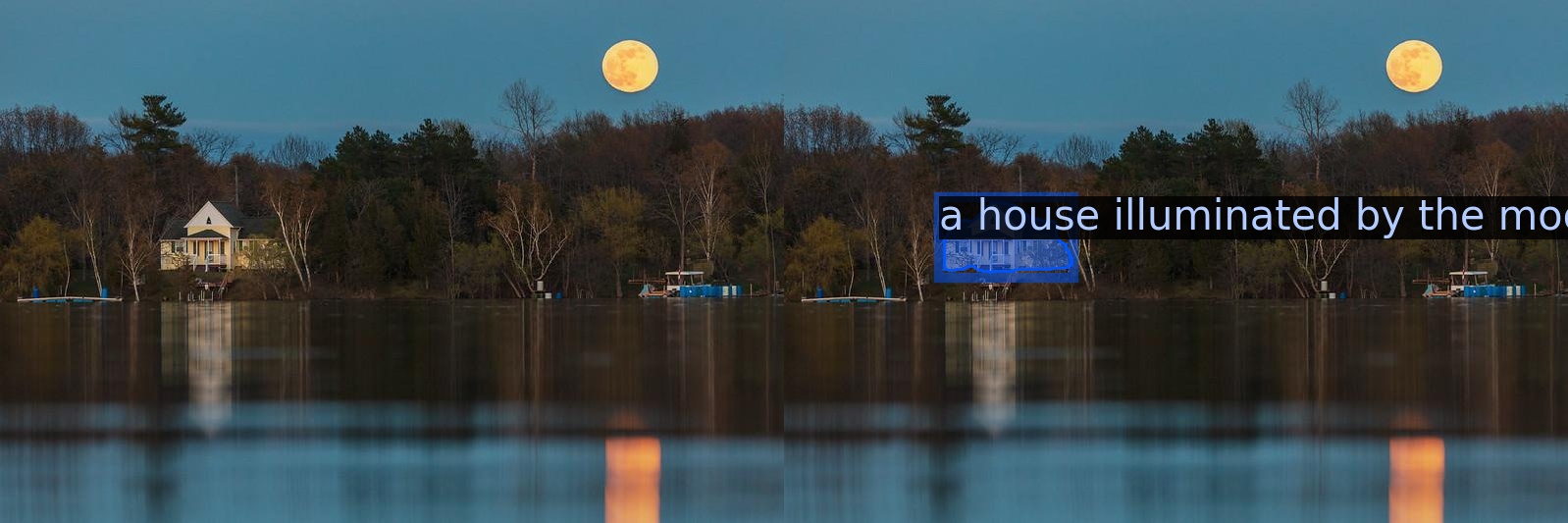}
		\caption{``a house illuminated by the moon''}
	\end{subfigure}
	
	\begin{subfigure}[t]{0.5\textwidth}
		\centering
		\includegraphics[height=1.2in]{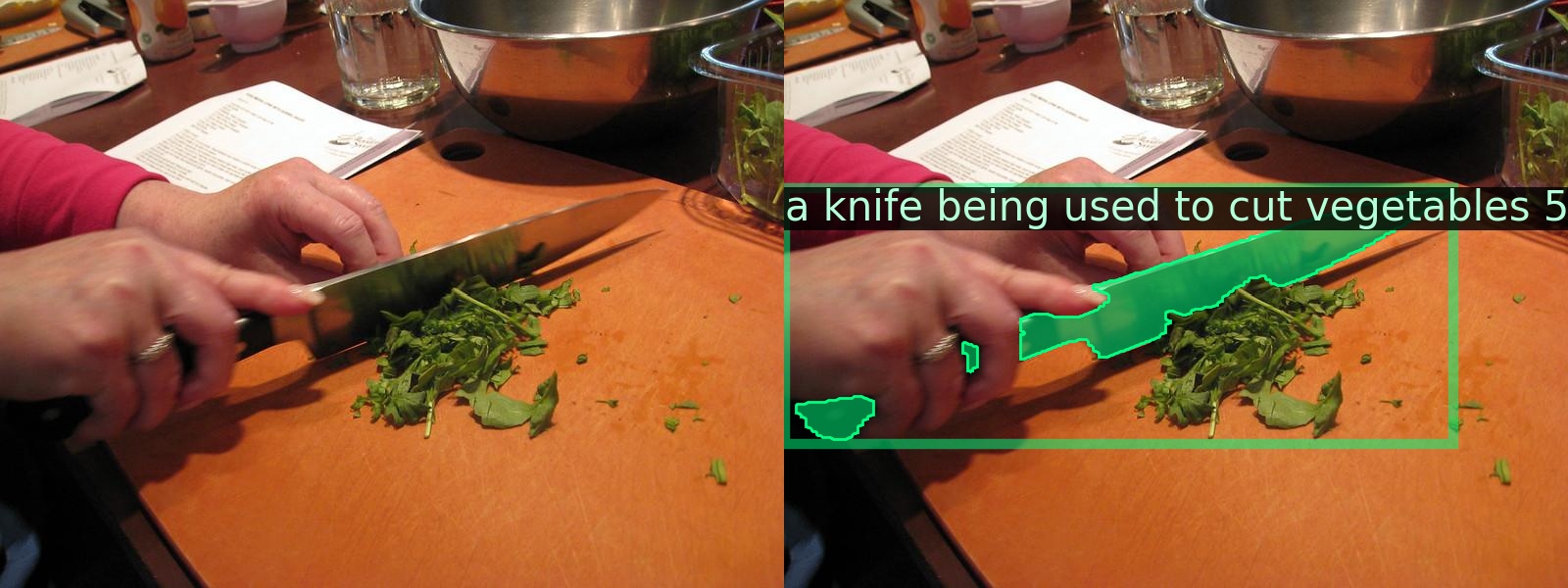}
		\caption{``a knife being used to cut vegetables''}
	\end{subfigure}%
	~ 
	\begin{subfigure}[t]{0.5\textwidth}
		\centering
		\includegraphics[height=1.2in]{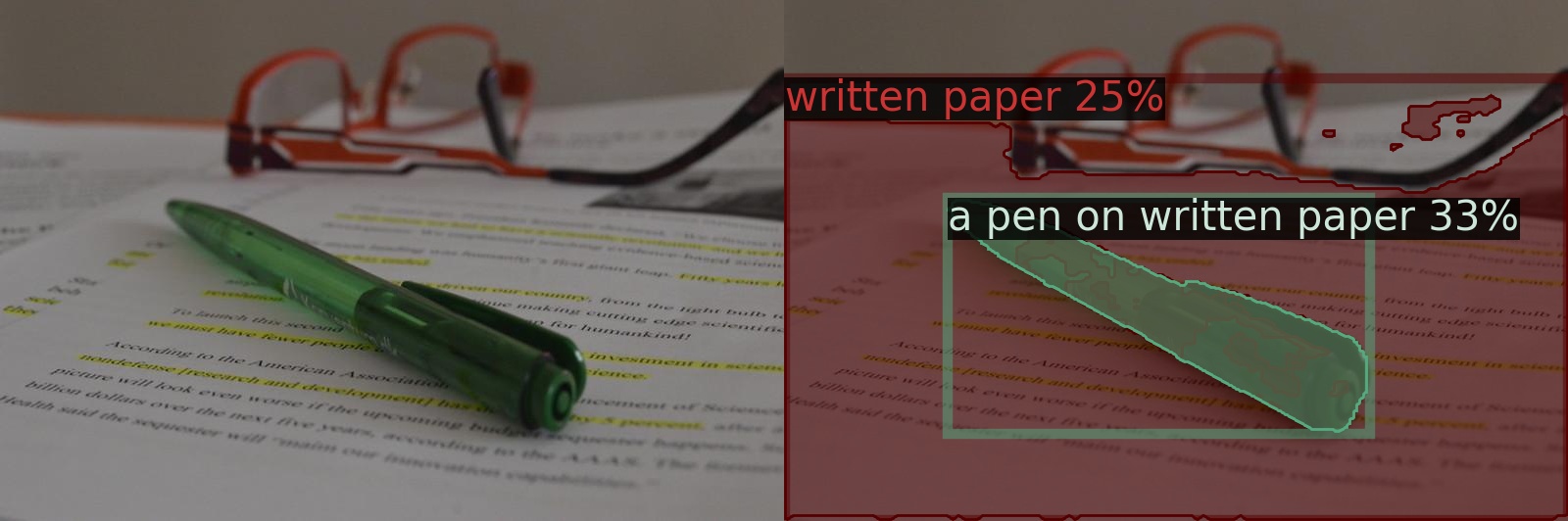}
		\caption{``written paper'', ``a pen on written paper''}
	\end{subfigure}
	
	\begin{subfigure}[t]{0.5\textwidth}
		\centering
		\includegraphics[height=1.2in]{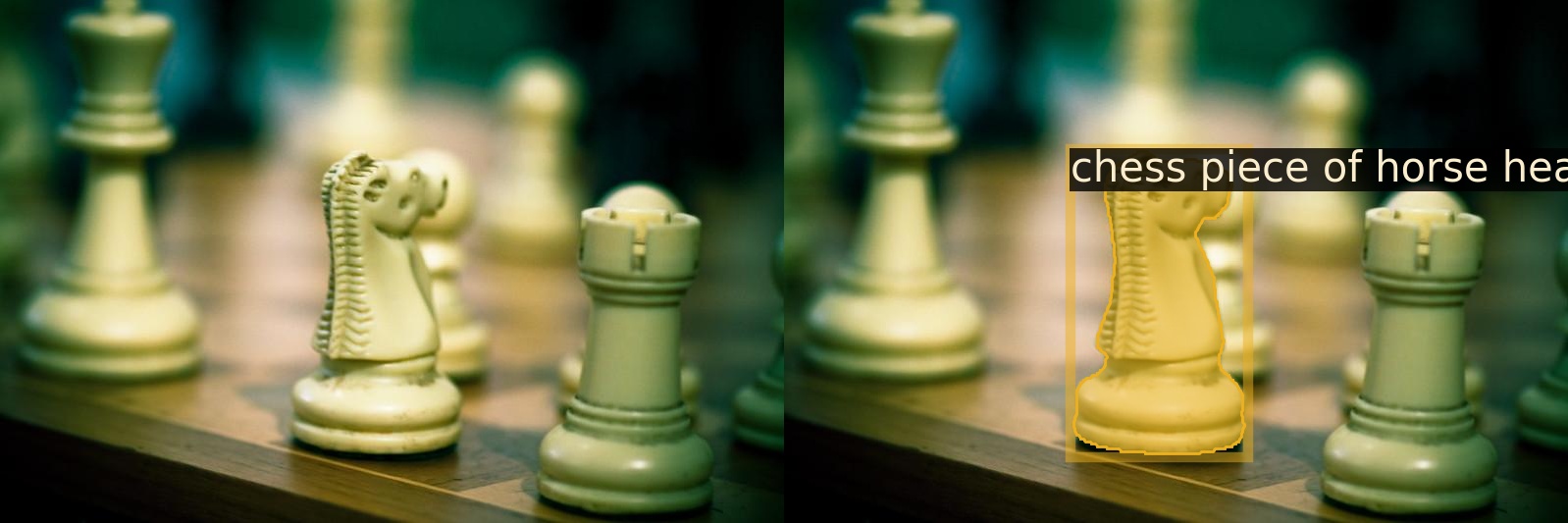}
		\caption{``chess piece of horse head''}
	\end{subfigure}%
	~ 
	\begin{subfigure}[t]{0.5\textwidth}
		\centering
		\includegraphics[height=1.2in]{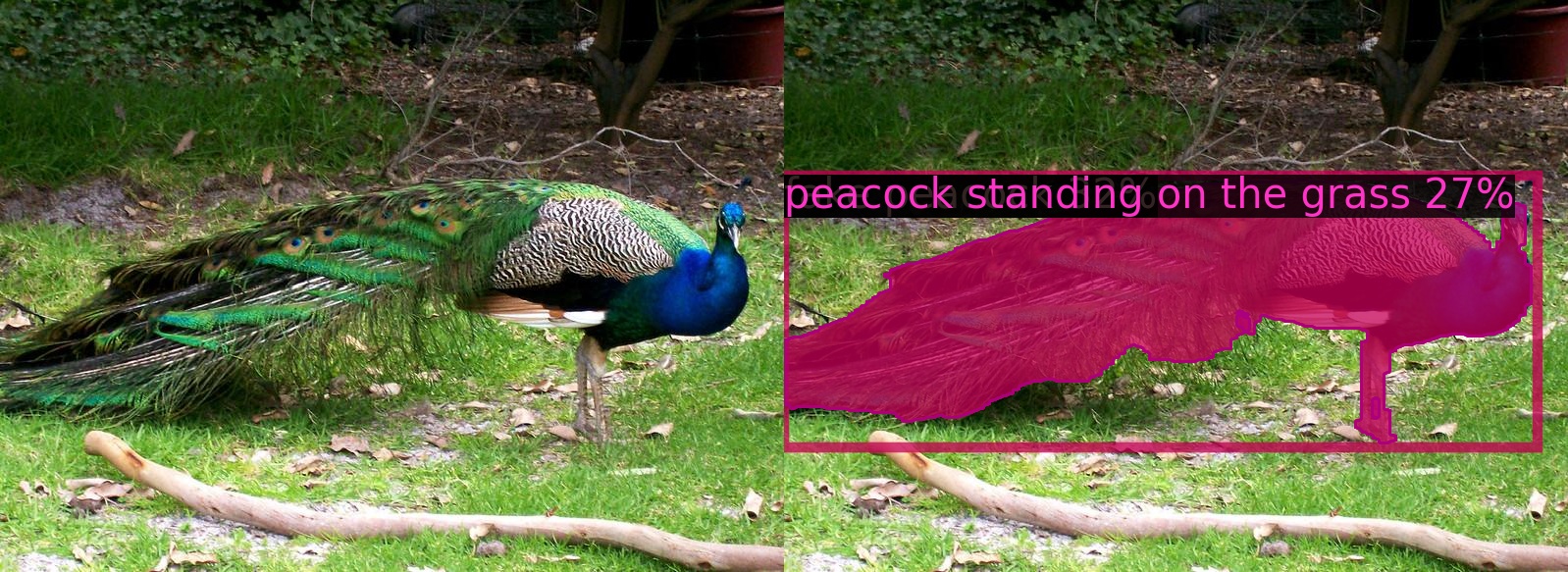}
		\caption{``peacock standing on the grass''}
	\end{subfigure}
	
	\begin{subfigure}[t]{1.0\textwidth}
		\centering
		\includegraphics[height=1.2in]{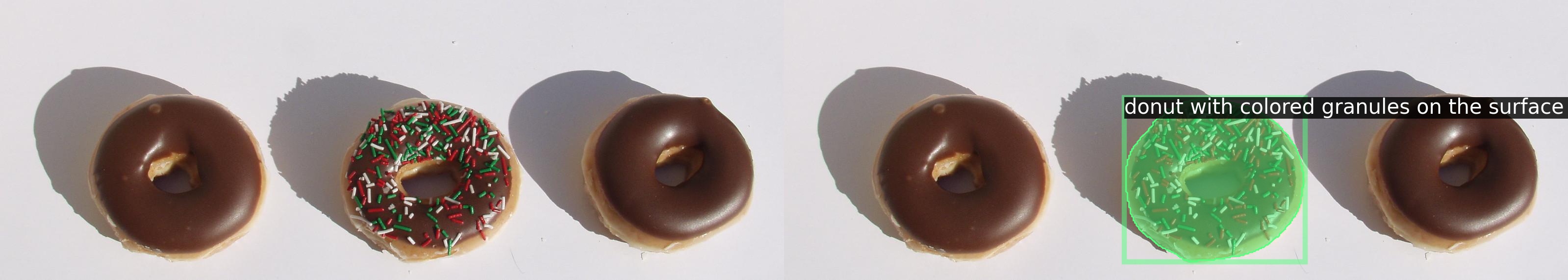}
		\caption{``donut with colored granules on the surface''}
	\end{subfigure}%
	
	\vspace{10pt}
	\caption{
		Visualizations of model outputs on D3~\cite{DOD}.
		In each group, the \textbf{left} image is the original image and the \textbf{right} image shows the predictions, and corresponding prompts of predicted objects are listed in the \textbf{subcaption}.
		All results are inferred in a single forward with all provide prompts.
	}
	\label{fig_vis_d3}
\end{figure*}

\begin{figure*}[t!]
	\centering

	\begin{subfigure}[t]{0.5\textwidth}
		\centering
		\includegraphics[height=1.2in]{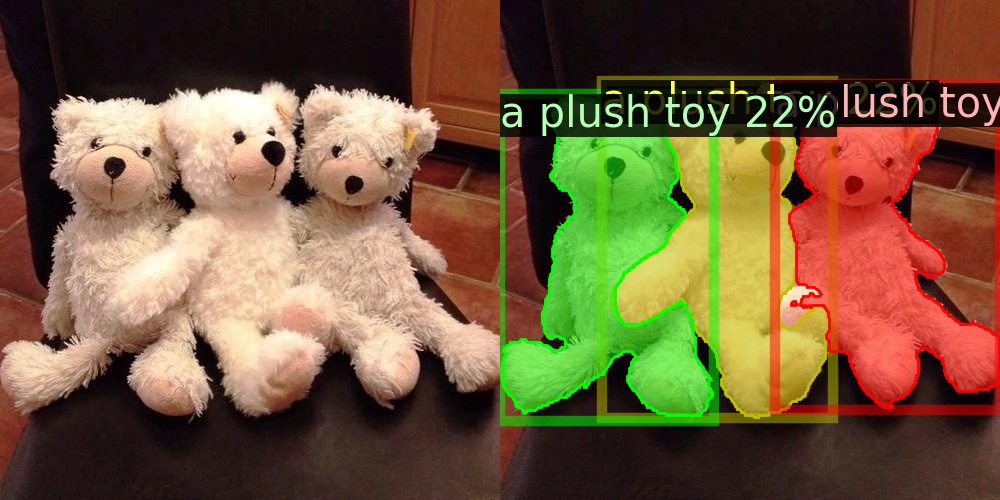}
		\caption{``a plush toy''}
	\end{subfigure}%
	~
	\begin{subfigure}[t]{0.5\textwidth}
		\centering
		\includegraphics[height=1.2in]{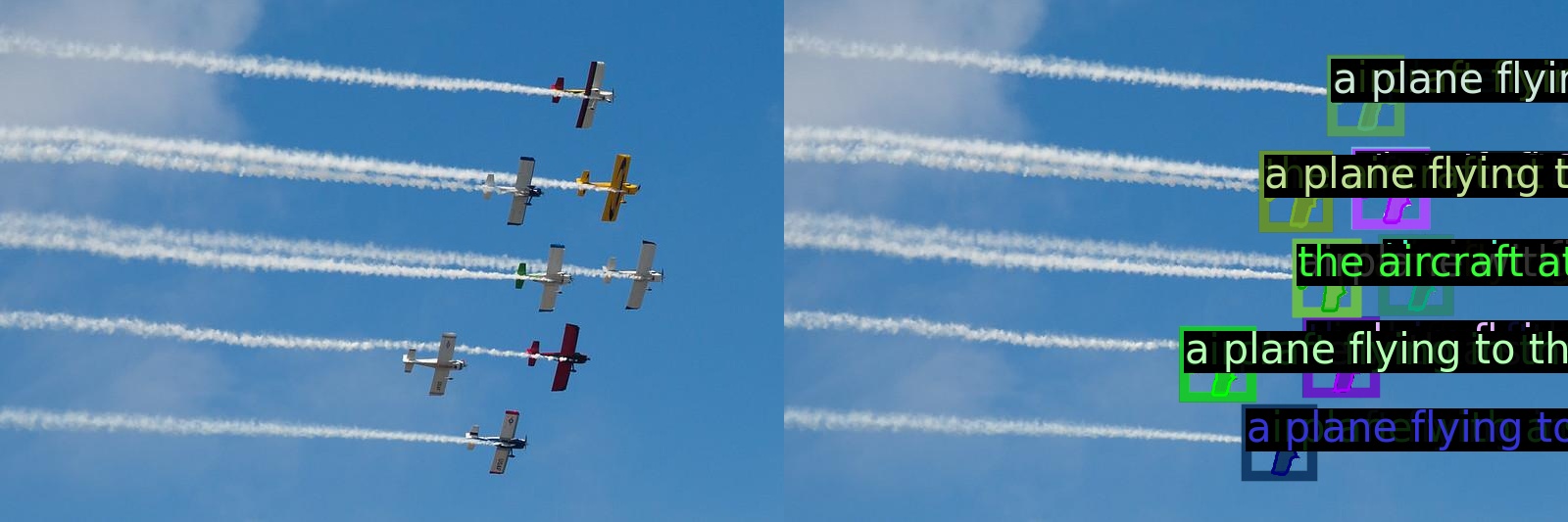}
		\caption{``a plane flying to the right''}
	\end{subfigure}
	
	\begin{subfigure}[t]{0.5\textwidth}
		\centering
		\includegraphics[height=1.2in]{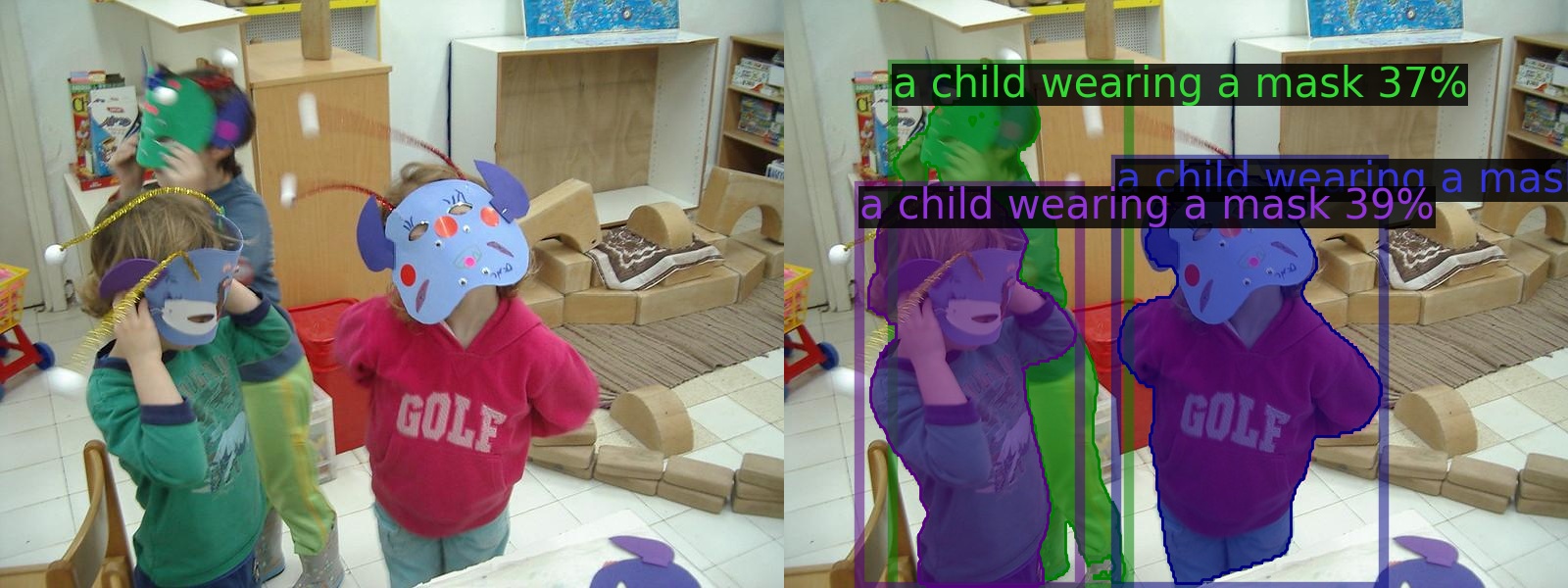}
		\caption{``a child wearing a mask''}
	\end{subfigure}%
	~ 
	\begin{subfigure}[t]{0.5\textwidth}
		\centering
		\includegraphics[height=1.2in]{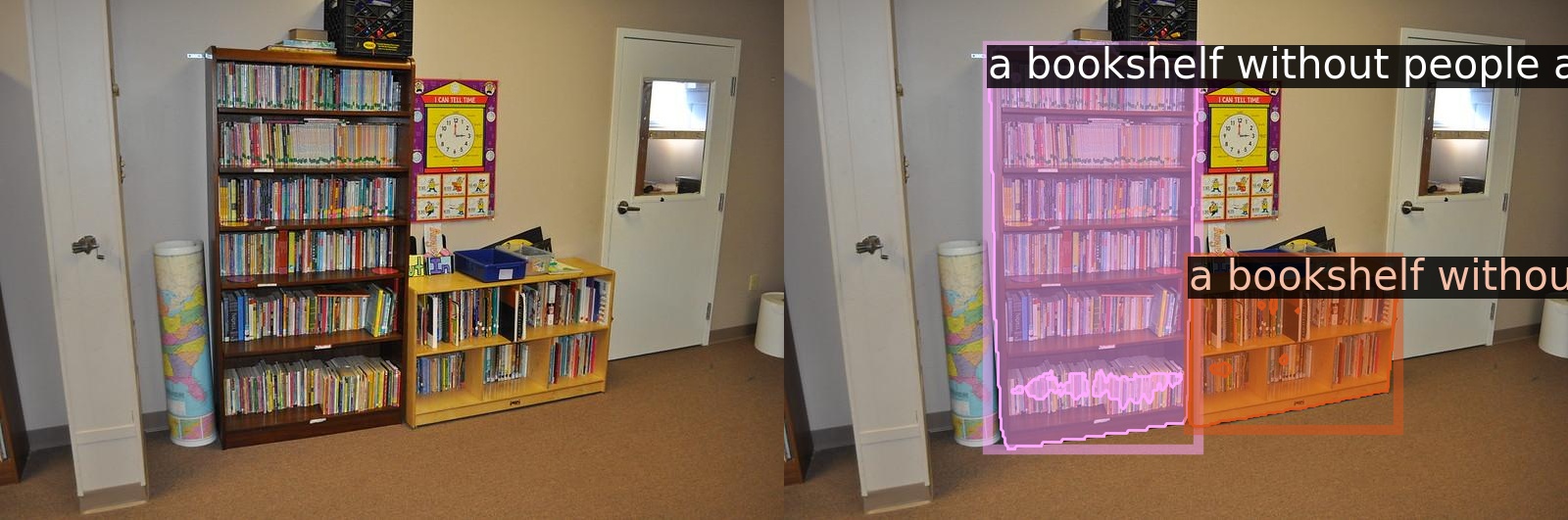}
		\caption{``a bookshelf without people around''}
	\end{subfigure}

	\begin{subfigure}[t]{0.3\textwidth}
		\centering
		\includegraphics[height=1.2in]{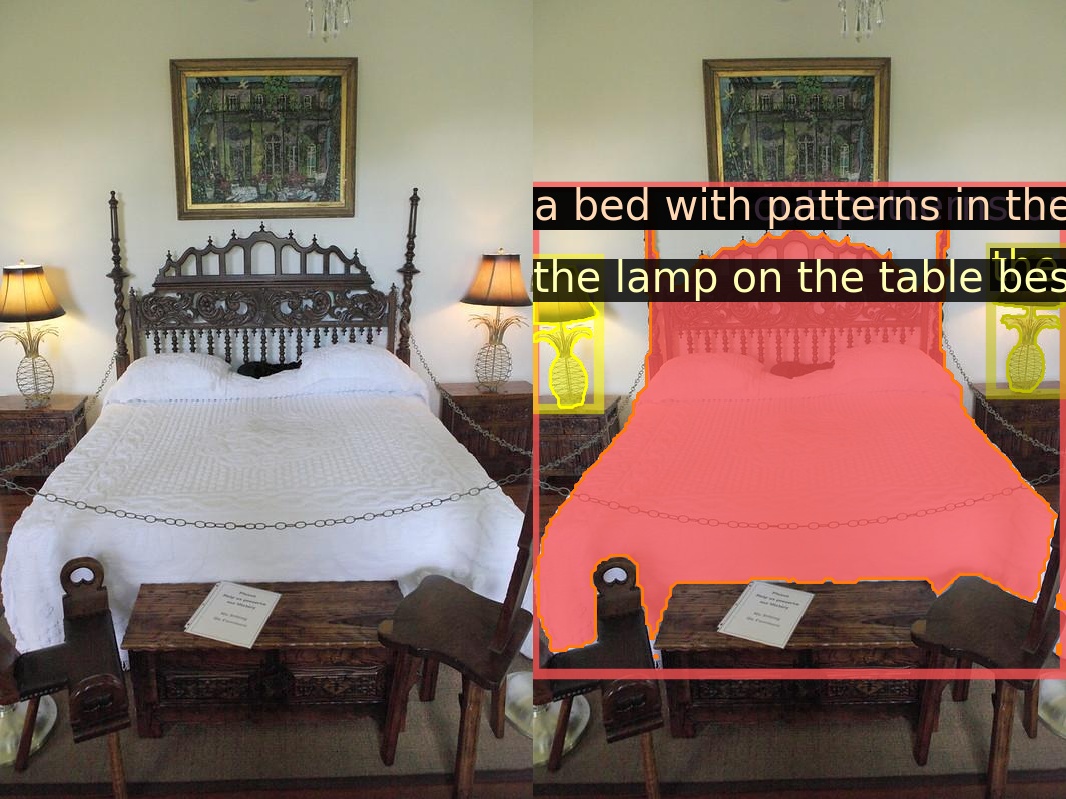}
		\caption{``a bed with patterns in the room'', ``the lamp on the table beside the bed''}
	\end{subfigure}%
	~ 
	\begin{subfigure}[t]{0.7\textwidth}
		\centering
		\includegraphics[height=1.2in]{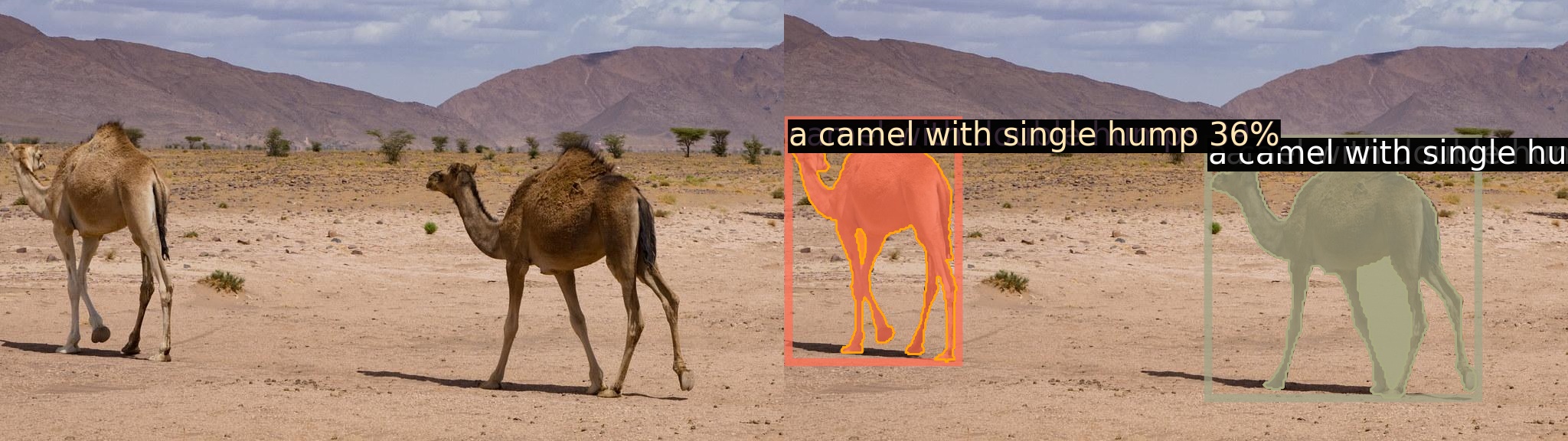}
		\caption{``a camel with single hump''}
	\end{subfigure}
	
	\vspace{10pt}
	\caption{
		Visualizations of model outputs on D3~\cite{DOD}.
		\OurMethod is capable to predict multiple instances for one sentence prompts.
		In each group, the \textbf{left} image is the original image and the \textbf{right} image shows the predictions, and corresponding prompts of predicted objects are listed in the \textbf{subcaption}.
		All results are inferred in a single forward with all provide prompts.
	}
	\label{fig_vis_d3_multiple}
\end{figure*}

\begin{figure*}[t!]
	\centering
	\begin{subfigure}[t]{0.6\textwidth}
		\centering
		\includegraphics[height=1.2in]{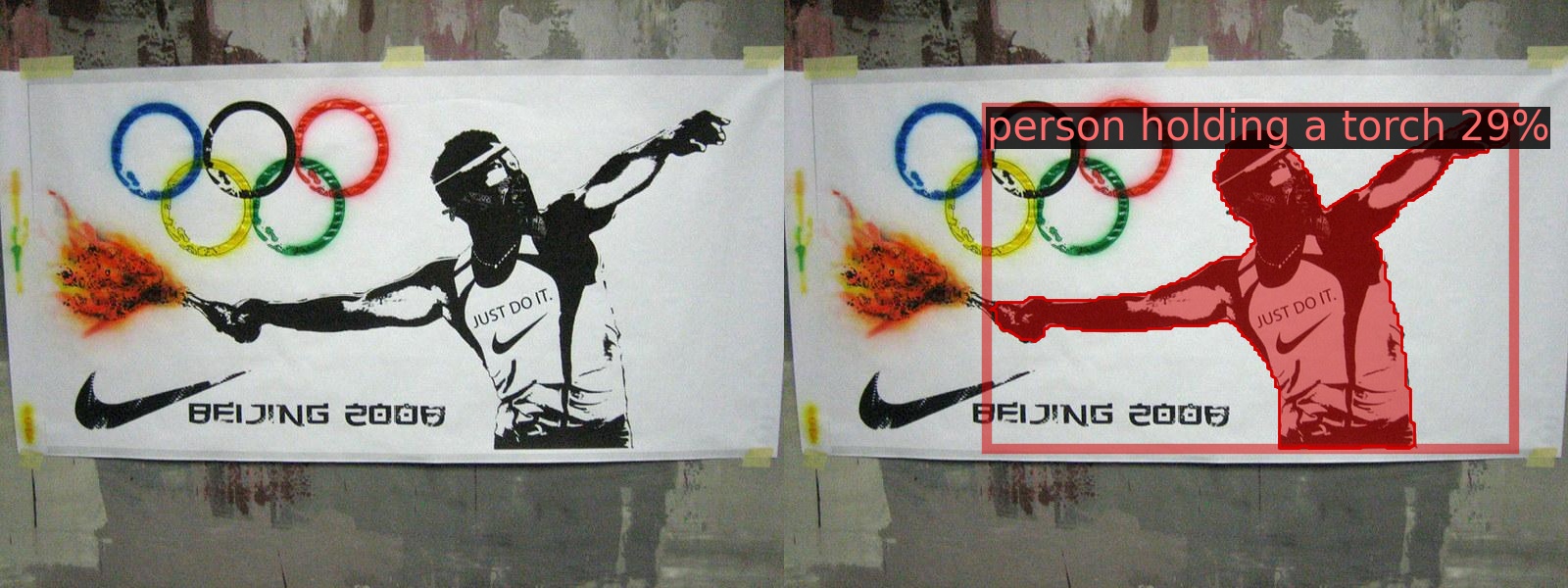}
		\caption{``person holding a torch''}
	\end{subfigure}%
	~
	\begin{subfigure}[t]{0.4\textwidth}
		\centering
		\includegraphics[height=1.2in]{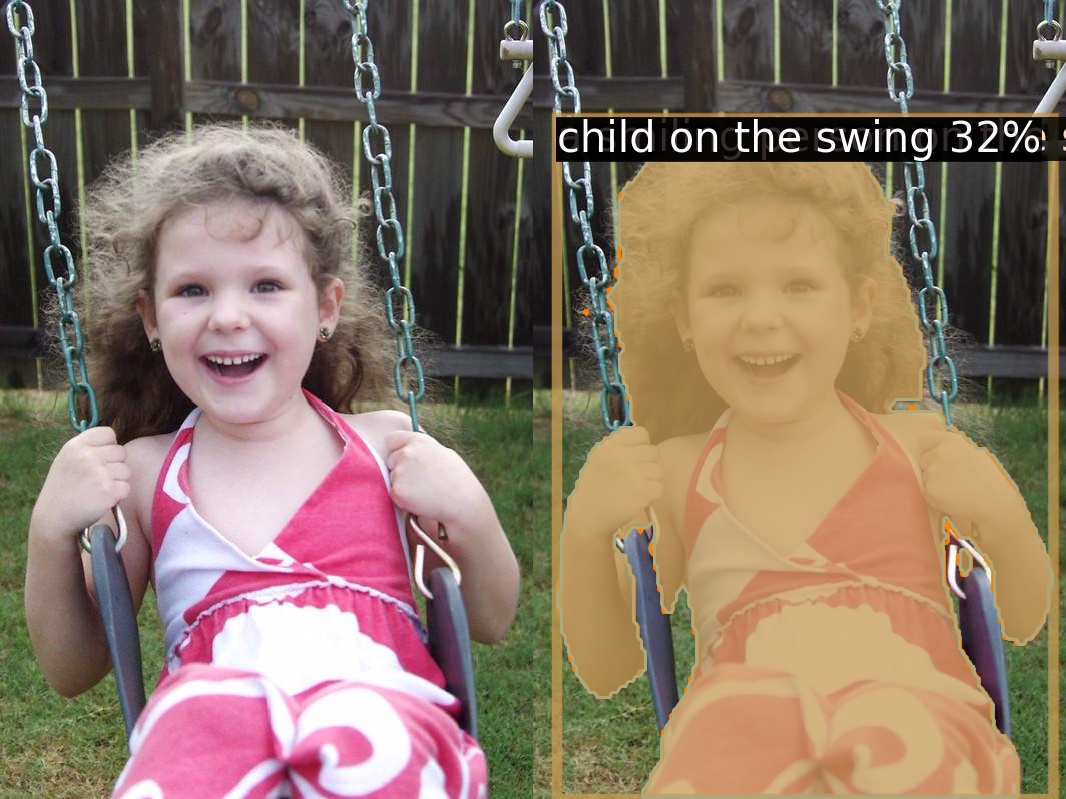}
		\caption{``child on the swing''}
	\end{subfigure}%

	\begin{subfigure}[t]{0.6\textwidth}
		\centering
		\includegraphics[height=1.2in]{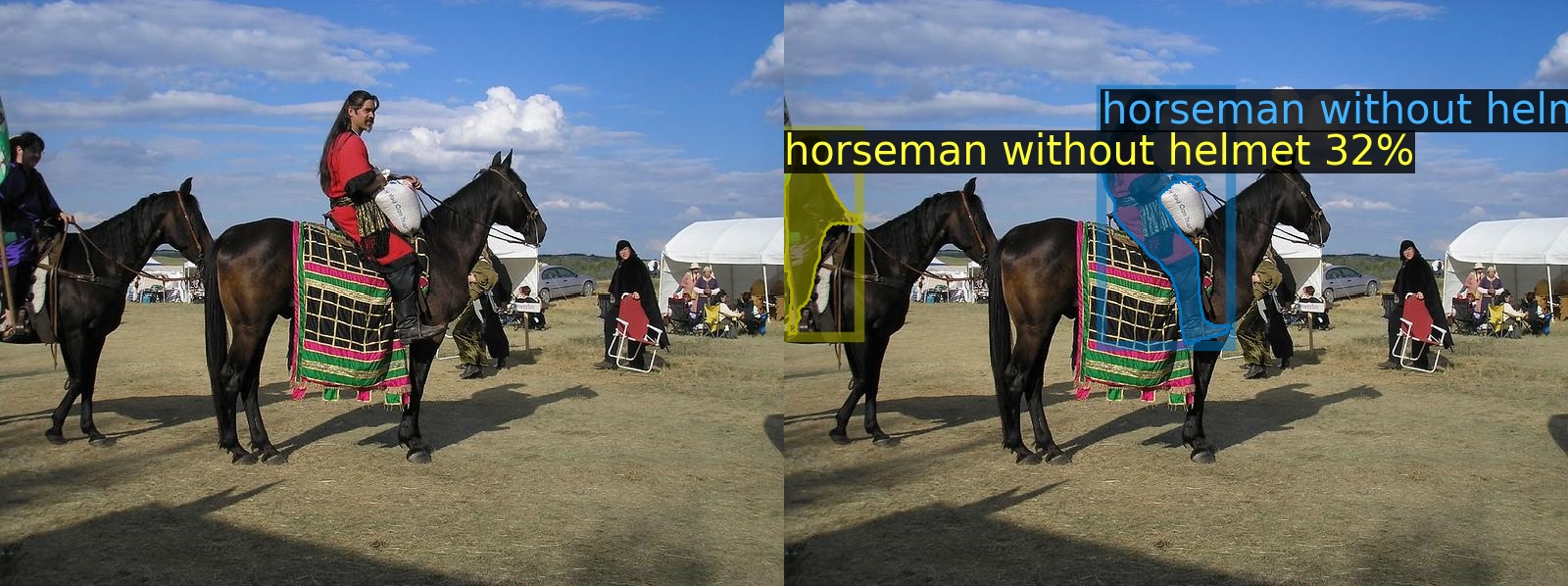}
		\caption{``horseman without helmet''}
	\end{subfigure}%
	~
	\begin{subfigure}[t]{0.4\textwidth}
		\centering
		\includegraphics[height=1.2in]{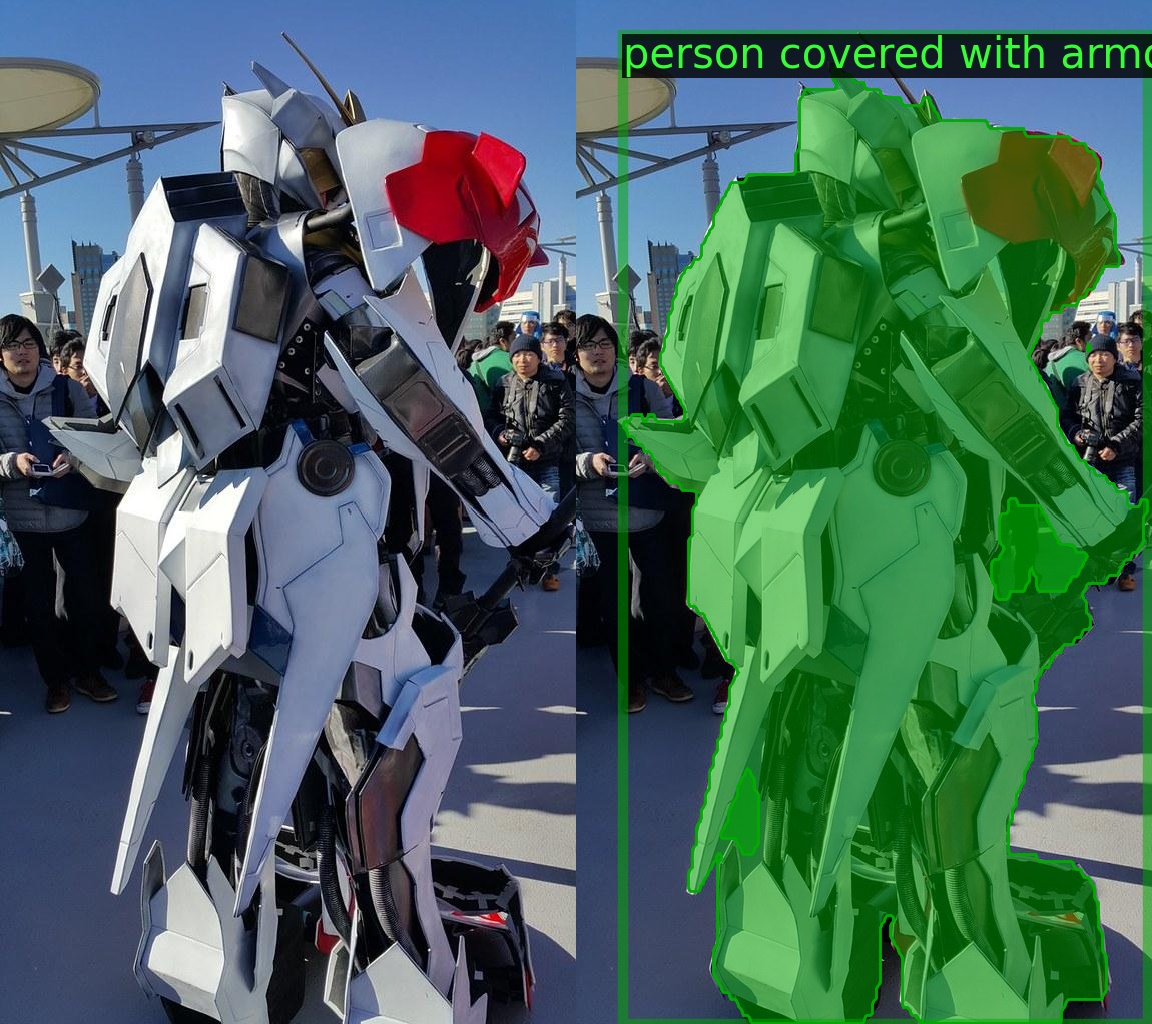}
		\caption{``person covered with armor''}
	\end{subfigure}
	
	\begin{subfigure}[t]{0.6\textwidth}
		\centering
		\includegraphics[height=1.2in]{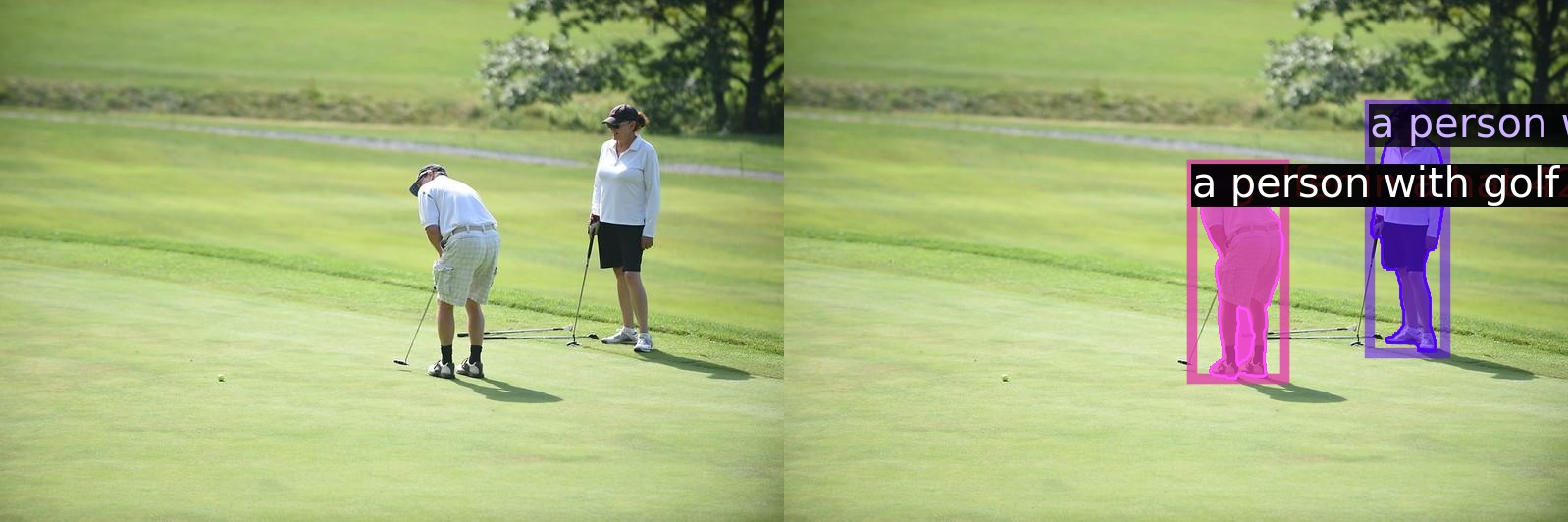}
		\caption{``a person with golf clubs''}
	\end{subfigure}%
	~
	\begin{subfigure}[t]{0.4\textwidth}
		\centering
		\includegraphics[height=1.2in]{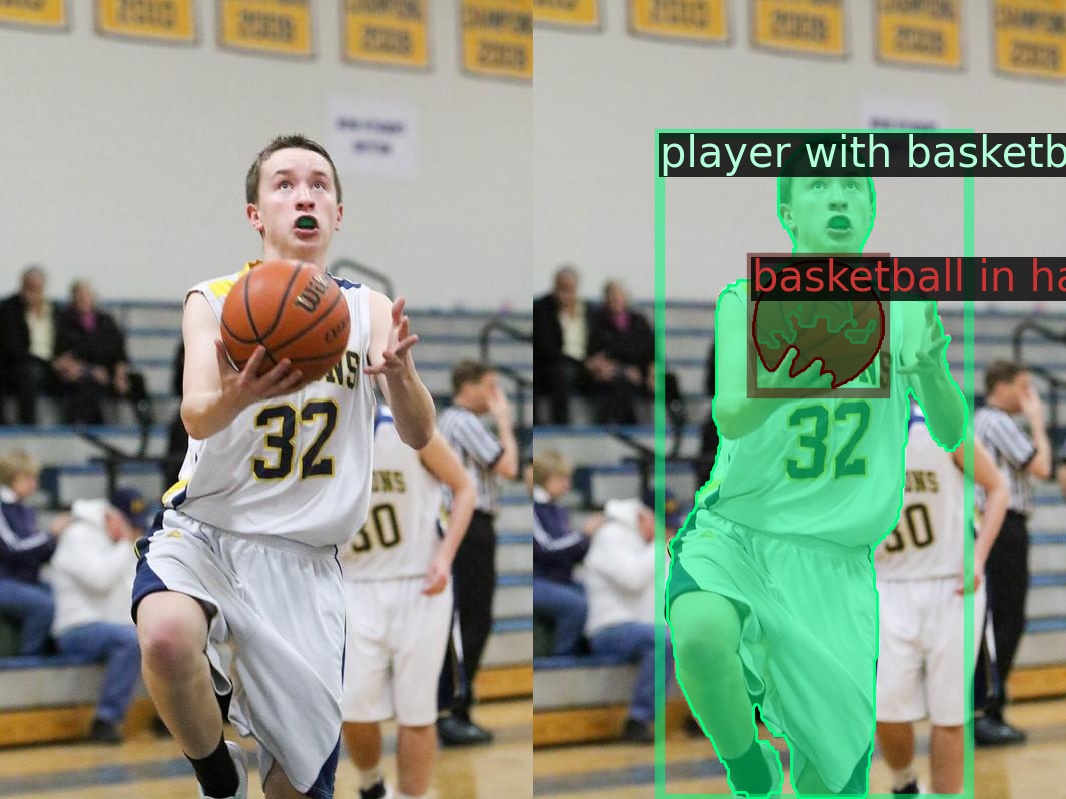}
		\caption{``player with basketball in the hand'', ``basketball in hand''}
	\end{subfigure}

	\begin{subfigure}[t]{0.6\textwidth}
		\centering
		\includegraphics[height=1.2in]{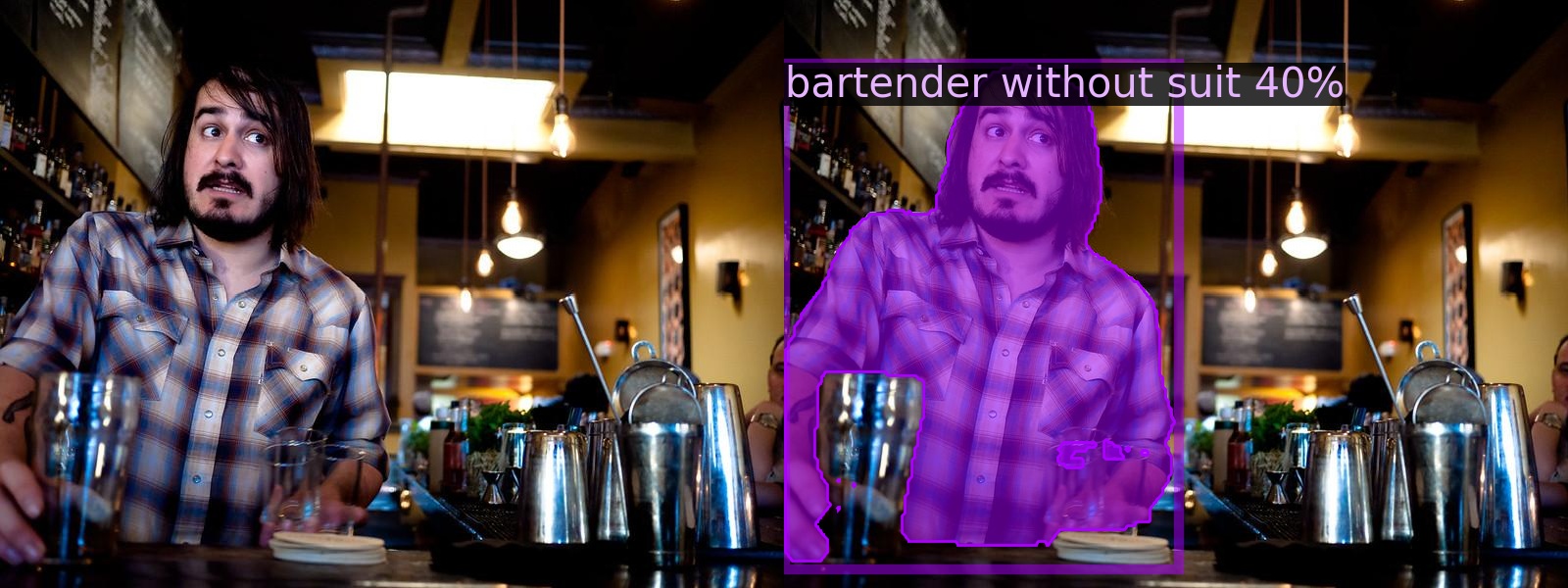}
		\caption{``bartender without suit''}
	\end{subfigure}%
	~
	\begin{subfigure}[t]{0.4\textwidth}
		\centering
		\includegraphics[height=1.2in]{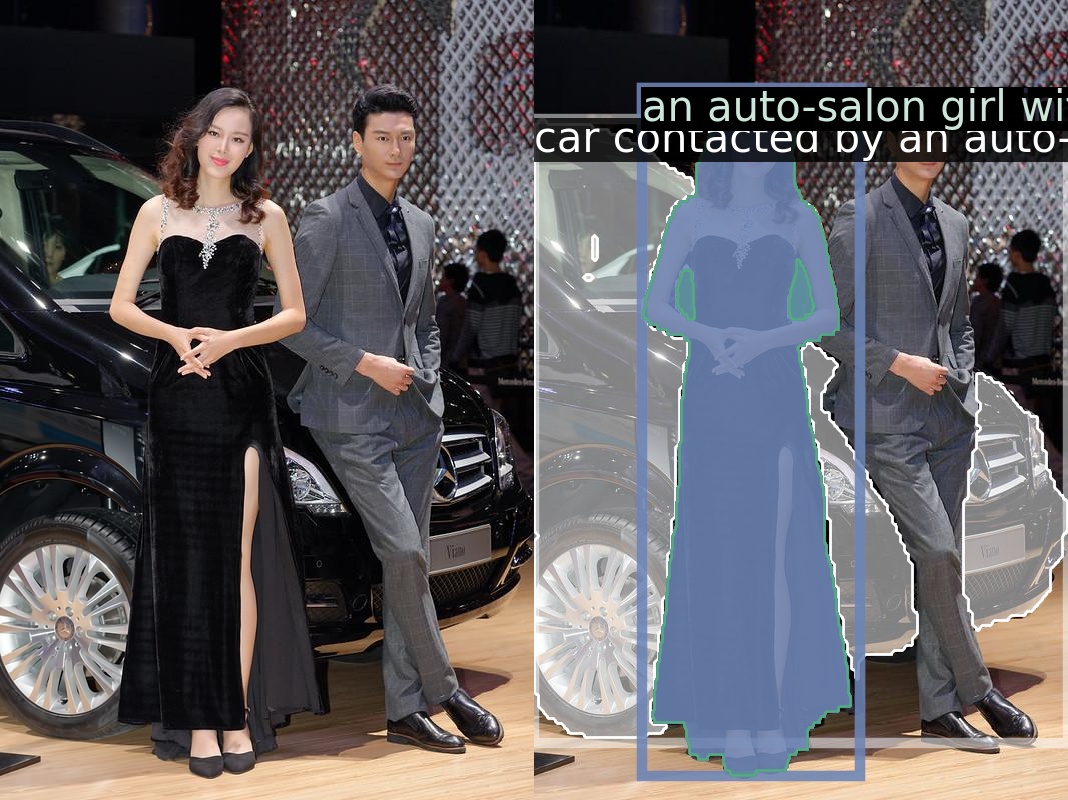}
		\caption{``car contacted by an auto-salon girl'', ``an auto-salon girl without bare waist''}
	\end{subfigure}

	\vspace{10pt}
	\caption{
		Visualizations of model outputs on D3~\cite{DOD} for Human-centric grounding.
		In each group, the \textbf{left} image is the original image and the \textbf{right} image shows the predictions, and corresponding prompts of predicted objects are listed in the \textbf{subcaption}.
		All results are inferred in a single forward with all provide prompts.
	}
	\label{fig_vis_d3_human}
\end{figure*}

We further present some visualizations in Figs.~\ref{fig_vis_d3}, ~\ref{fig_vis_d3_multiple} and ~\ref{fig_vis_d3_human} on D3~\cite{DOD}, on which \OurMethod outperforms all previous methods with a large gap.
Our \OurMethod presents great generalization on different scenes and text inputs.

\begin{figure*}[t!]
	\centering
	
	\begin{subfigure}[t]{0.5\textwidth}
		\centering
		\includegraphics[height=1.2in]{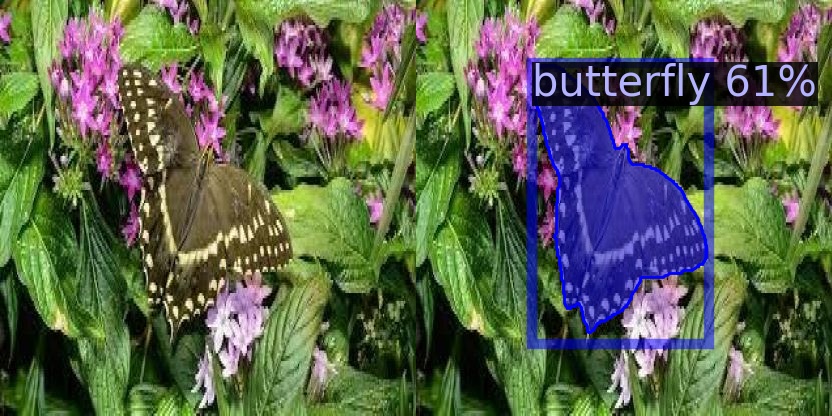}
		\caption{``butterfly''}
	\end{subfigure}%
	~ 
	\begin{subfigure}[t]{0.5\textwidth}
		\centering
		\includegraphics[height=1.2in]{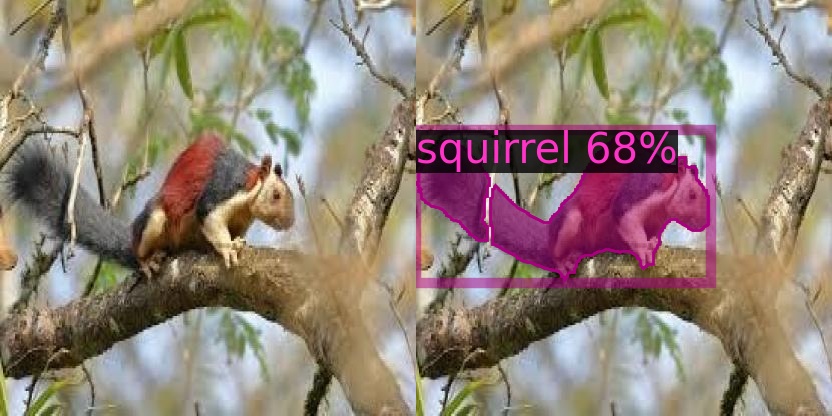}
		\caption{``squirrel''}
	\end{subfigure}
	
	\begin{subfigure}[t]{0.5\textwidth}
		\centering
		\includegraphics[height=1.2in]{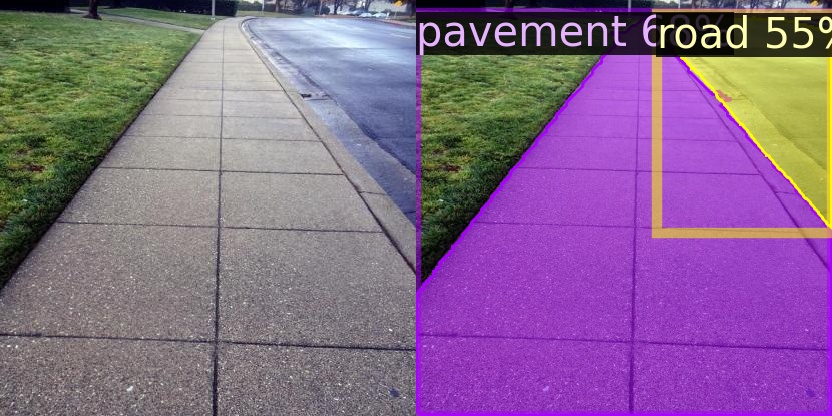}
		\caption{``pavement'', ``road''}
	\end{subfigure}%
	~
	\begin{subfigure}[t]{0.5\textwidth}
		\centering
		\includegraphics[height=1.2in]{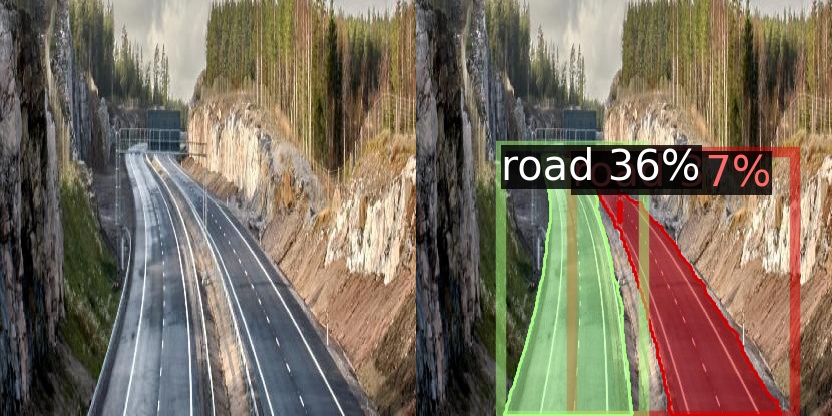}
		\caption{``road''}
	\end{subfigure}
	
	\begin{subfigure}[t]{0.5\textwidth}
		\centering
		\includegraphics[height=1.2in]{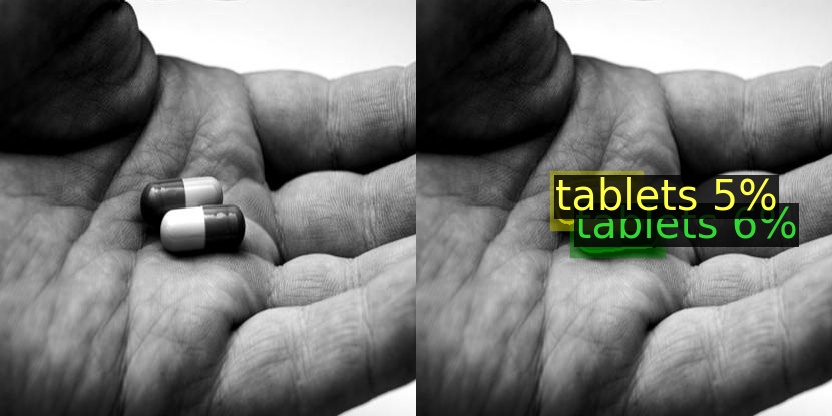}
		\caption{``tablets''}
	\end{subfigure}%
	~ 
	\begin{subfigure}[t]{0.5\textwidth}
		\centering
		\includegraphics[height=1.2in]{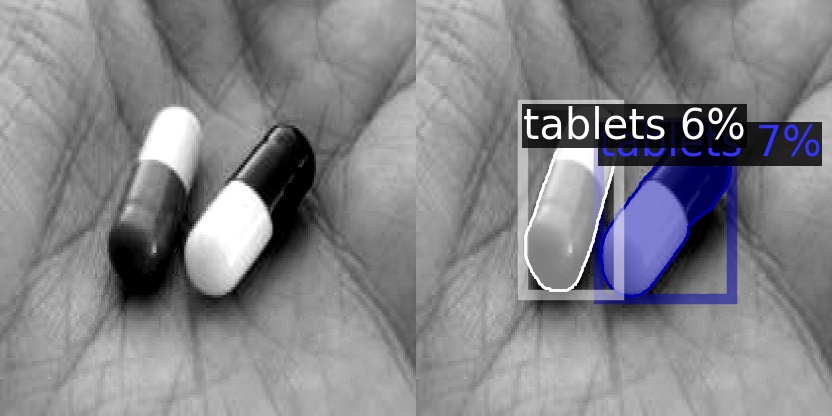}
		\caption{``tablets''}
	\end{subfigure}
	
	\begin{subfigure}[t]{0.5\textwidth}
		\centering
		\includegraphics[height=1.2in]{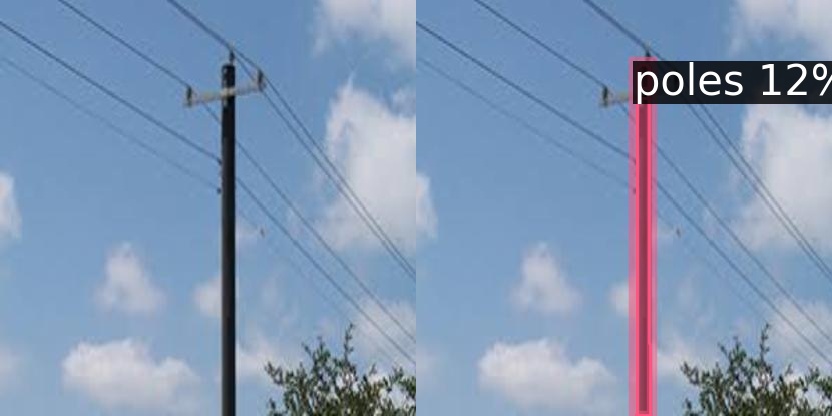}
		\caption{``poles''}
	\end{subfigure}%
	~ 
	\begin{subfigure}[t]{0.5\textwidth}
		\centering
		\includegraphics[height=1.2in]{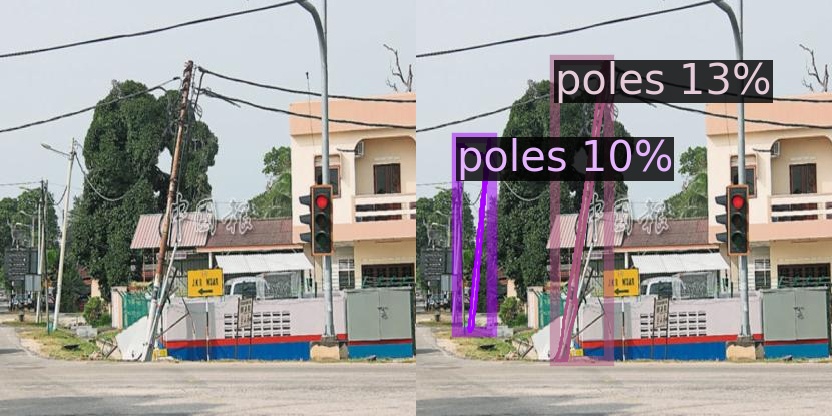}
		\caption{``poles''}
	\end{subfigure}
	
	\vspace{10pt}
	\caption{
		Visualizations of model outputs on SegInW~\cite{X-Decoder}.
		In each group, the \textbf{left} image is the original image and the \textbf{right} image shows the predictions, and corresponding prompts of predicted objects are listed in the \textbf{subcaption}.
		All results are inferred in a single forward with all provide prompts.
	}
	\label{fig_vis_seginw}
\end{figure*}

Finally, we visualize some examples on SegInW~\cite{X-Decoder} in Fig.~\ref{fig_vis_seginw}.

\end{document}